\documentclass{article}
\usepackage{iclr2021_conference,times}

\usepackage{hyperref}
\usepackage{url}

\usepackage{amsmath,amsfonts,bm,physics}

\newcommand{\Unif}{{\mathrm{Unif}}}

\def\mTheta{{\bm{\Theta}}}

\def\hatvtheta{{\widehat{\vtheta}}}
\def\hatmTheta{{\widehat{\mTheta}}}
\def\hatvw{{\widehat{\vw}}}
\def\hatvb{{\widehat{\vb}}}
\def\hatmB{{\widehat{\mB}}}
\def\hatmW{{\widehat{\mW}}}
\def\hatmM{{\widehat{\mM}}}

\DeclareMathOperator{\ind}{\mathbb{I}}

\newcommand{\mat}[1]{\mathbf{#1}}

\def\eqref#1{equation~\ref{#1}}

\def\1{\bm{1}}

\def\vzero{{\bm{0}}}

\def\vtheta{{\bm{\theta}}}
\def\va{{\bm{a}}}
\def\vb{{\bm{b}}}

\def\ve{{\bm{e}}}

\def\vu{{\bm{u}}}
\def\vv{{\bm{v}}}
\def\vw{{\bm{w}}}
\def\vx{{\bm{x}}}
\def\vy{{\bm{y}}}

\def\mA{{\bm{A}}}
\def\mB{{\bm{B}}}

\def\mD{{\bm{D}}}

\def\mI{{\bm{I}}}

\def\mM{{\bm{M}}}

\def\mQ{{\bm{Q}}}

\def\mW{{\bm{W}}}
\def\mX{{\bm{X}}}

\DeclareMathAlphabet{\mathsfit}{\encodingdefault}{\sfdefault}{m}{sl}
\SetMathAlphabet{\mathsfit}{bold}{\encodingdefault}{\sfdefault}{bx}{n}

\def\gA{{\mathcal{A}}}
\def\gB{{\mathcal{B}}}

\def\gD{{\mathcal{D}}}

\def\gG{{\mathcal{G}}}

\def\gI{{\mathcal{I}}}

\def\gN{{\mathcal{N}}}

\def\gS{{\mathcal{S}}}

\def\gW{{\mathcal{W}}}

\def\sR{{\mathbb{R}}}
\def\sS{{\mathbb{S}}}

\newcommand{\E}{\mathbb{E}}

\DeclareMathOperator*{\argmax}{arg\,max}
\DeclareMathOperator*{\argmin}{arg\,min}

\usepackage{hyperref}

\usepackage{url}
\usepackage{amsthm, amssymb}
\usepackage{verbatim}
\usepackage[inline]{enumitem}
\usepackage{xcolor}
\usepackage[ruled,noline,noend,linesnumbered]{algorithm2e}
\usepackage{pgf}

\hypersetup{
  colorlinks   = true, 
  urlcolor     = blue, 
  linkcolor    = blue, 
  citecolor   = blue 
}
\setlist[enumerate, 1]{label=(\alph*)}
\setlist[enumerate]{nosep}

\theoremstyle{plain}
\newtheorem{thm}{Theorem}
\newtheorem{lem}{Lemma}
\newtheorem{coro}[thm]{Corollary}

\theoremstyle{definition}

\newtheorem{assum}{Assumption}

\theoremstyle{remark}

\SetCommentSty{rmfamily}

\SetKwComment{Comment}{$\triangleright$\ }{}

\renewcommand{\eqref}[1]{\hyperref[#1]{(\ref{#1})}}

\SetNlSty{textbf}{}{:}

\renewcommand{\E}{\mathop{\mathbb E}}

\newcommand{\algoname}{$\mathsf{E^2TC}$}

\newcommand{\red}[1]{#1}

\title{\red{Impact} of Representation Learning in Linear Bandits}

\iclrfinalcopy

\author{Jiaqi Yang \\
Tsinghua University\\
\texttt{yangjq17@gmail.com} \\
\And
Wei Hu \\
Princeton University \\
\texttt{huwei@cs.princeton.edu} \\
\AND
Jason D. Lee \\
Princeton University \\
\texttt{jasonlee@princeton.edu}
\And
Simon S. Du \\ 
University of Washington \\
\texttt{ssdu@cs.washington.edu}
}

\begin{document}

\maketitle

\begin{abstract}
We study how representation learning can improve the efficiency of bandit problems. 
We study the setting where we play $T$ linear bandits with dimension $d$ concurrently, and these $T$ bandit tasks share a common $k (\ll d)$ dimensional linear representation.
For the finite-action setting, we present a new algorithm which achieves $\widetilde{O}(T\sqrt{kN} + \sqrt{dkNT})$ regret, where $N$ is the number of rounds we play for each bandit. When $T$ is sufficiently large, our algorithm significantly outperforms the naive algorithm (playing $T$ bandits independently) that achieves $\widetilde{O}(T\sqrt{d N})$ regret. We also provide an $\Omega(T\sqrt{kN} + \sqrt{dkNT})$ regret lower bound, showing that our algorithm is minimax-optimal up to poly-logarithmic factors. 
Furthermore, we extend our algorithm to the infinite-action setting and obtain a corresponding regret bound which demonstrates the benefit of representation learning in certain regimes.
We also present experiments on synthetic and real-world data to illustrate our theoretical findings and demonstrate the effectiveness of our proposed algorithms.
\end{abstract}

\section{Introduction}
\label{sec:intro}
This paper investigates the benefit of using representation learning for sequential decision-making problems.
Representation learning learns a joint low-dimensional embedding (feature extractor) from different but related tasks and then uses a simple function (often a linear one) on top of the embedding \citep{baxter2000model,caruana1997multitask,li2010contextual}
The mechanism behind is that since the tasks are related, we can extract the common information more efficiently than treating each task independently.

Empirically, representation learning has become a popular approach for improving sample efficiency across various machine learning tasks~\citep{bengio2013representation}.
In particular, recently, representation learning has become increasingly more popular in sequential decision-making problems \citep{teh2017distral,taylor2009transfer,lazaric2011transfer,rusu2015policy,liu2016decoding,parisotto2015actor,higgins2017darla,hessel2019multi,arora2020provable,d'eramo2020sharing}.
 For example, many sequential decision-making tasks share the same environment but have different reward functions.
Thus a natural approach is to learn a succinct representation that describes the environment and then make decisions for different tasks on top of the learned representation.

While representation learning is already widely applied in sequential decision-making problems empirically, its theoretical foundation is still limited.
One important problem remains open:
\begin{center}
\textbf{When does representation learning \emph{provably} improve efficiency of sequential decision-making problems?}
\end{center}
%
\red{We} take a step to characterize the benefit of representation learning in sequential decision-making problems.
We tackle the above problem in the linear bandits setting, one of the most fundamental and popular settings in sequential decision-making problems.
This model is widely used in applications as such clinical treatment, manufacturing process, job scheduling, recommendation systems, etc~\citep{dani2008stochastic,chu2011contextual}.

\red{We} study the multi-task version of linear bandits, which naturally models the scenario where one needs to deal with multiple different but closely related sequential decision-making problems concurrently.

We will mostly focus on the finite-action setting. Specifically, we have $T$ tasks, each of which is governed by an unknown linear coefficient $\vtheta_t \in \mathbb R^d$. At the $n$-th round, for each task $t \in [T]$, the player chooses an action $a_{n,t}$ that belongs to a finite set, and receive a reward $r_{n, t}$ with expectation $\E{r_{n, t}} = \langle\vtheta_t, \vx_{n,t, a_{n, t}}\rangle$ where $\vx_{n,t, a_{n, t}}$ represents the context of action $a_{n,t}$.
For this problem, a straightforward approach is to treat each task independently, which leads to  $\widetilde{O}(T\sqrt{dN})$\footnote{$\widetilde{O}(\cdot)$ omits logarithmic factors.} total regret.
Can we do better?

Clearly, if the tasks are independent, then by the classical $\Omega( \sqrt{dN})$ per task lower bound for linear bandit, it is impossible to do better. We investigate how representation learning can help if the tasks are related. Our main assumption is the existence of an unknown \emph{linear feature extractor} $\mB \in \sR^{d \times k}$ with $k \ll d$ and a set of linear coefficients $\{\vw_t\}_{t=1}^T$ such that $\vtheta_t = \mB \vw_t$. Under this assumption, the tasks are closely related as $\mB$ is a \emph{shared} linear feature extractor that maps the raw contexts $\vx_{n, t, a} \in \sR^d$ to a low-dimensional embedding $\mB^\top \vx_{n, t, a} \in \sR^k$. 
\red{In this paper, we focus on the regime where $k \ll d, N, T$.
This regime is common in real-world problems, e.g., computer vision, where the input dimension is high, the number of data is large, many task are related, and there exists a low-dimension representation among these tasks that we can utilize.
}
Problems with similar assumptions have been studied in the supervised learning setting \citep{ando2005framework}.
However, to our knowledge, this formulation has not been studied in the bandit setting.

\paragraph{Our Contributions}
We give the first rigorous characterization on the benefit of representation learning for multi-task linear bandits. Our contributions are summarized below.
\begin{itemize}[leftmargin=*]
	\item We design a new algorithm for the aforementioned problem. 
	Theoretically, we show our algorithm incurs $\widetilde{O}(\sqrt{dkTN} +  T \sqrt{kN})$ total regret in $N$ rounds for all $T$ tasks.
	Therefore, our algorithm outperforms the naive approach with $O(T\sqrt{dN})$ regret.
	To our knowledge, this is the first theoretical result demonstrating the benefit of representation learning for bandits problems.
	\item To complement our upper bound, we also provide an $\Omega(\sqrt{d k T N} +  T \sqrt{kN})$ lower bound, showing our regret bound is tight up to polylogarithmic factors.
	\item We further design a new algorithm for the infinite-action setting, which has a regret $\widetilde{O}(d^{1.5}k\sqrt{TN} + kT\sqrt{N})$, which outperforms the naive approach with $O(Td\sqrt{N})$ regret in the regime where $T = \widetilde{\Omega}(dk^2)$.
	\item We provide simulations and an experiment on MNIST dataset to illustrate the effectiveness of our algorithms and the benefits of representation learning.
\end{itemize}

\paragraph{Organization} This paper is organized as follows.
In \autoref{sec:rel}, we discuss related work.
In \autoref{sec:pre}, we introduce necessary notation, formally set up our problem, and describe our assumptions.
In \autoref{sec:upper_bound}, we present our main algorithm for the finite-action setting and its performance guarantee.
In \autoref{sec:infarm}, we describe our algorithm and its theoretical guarantee for the infinite-action setting.
In \autoref{sec:exp}, we provide simulation studies and real-world experiments to validate the effectiveness of our approach.
We conclude in \autoref{sec:conclusion} and defer all proofs to the Appendix.

\section{Related Work}
\label{sec:rel}
Here we mainly focus on related theoretical results. We refer readers to \citet{bengio2013representation} for empirical results of using representation learning.

For supervised learning, there is a long line of works on multi-task learning and representation learning with various assumptions \citep{baxter2000model,ando2005framework,ben2003exploiting,maurer2006bounds,cavallanti2010linear,maurer2016benefit,du2020few,tripuraneni2020provable}.
All these results assumed the existence of a common representation shared among all tasks.
However, this assumption alone is not sufficient.
For example, \citet{maurer2016benefit} further assumed every task is i.i.d. drawn from an underlying distribution.
Recently, \citet{du2020few} replaced the i.i.d. assumption with a deterministic assumption on the input distribution.
Finally, it is worth mentioning that \citet{tripuraneni2020provable} gave the method-of-moments estimator and built the confidence ball for the feature extractor, which inspired our algorithm for the infinite-action setting.

The benefit of representation learning has been studied in sequential decision-making problems, especially in reinforcement learning domains.
\citet{d'eramo2020sharing} showed that representation learning can improve the rate of approximate value iteration algorithm.
\citet{arora2020provable} proved that representation learning can reduce the sample complexity of imitation learning.
Both works require a probabilistic assumption similar to that in \citep{maurer2016benefit} and the statistical rates are of similar forms as those in \citep{maurer2016benefit}.

We remark that representation learning is also closely connected to meta-learning \citep{schaul2010metalearning}.
\citet{raghu2019rapid} empirically suggested that the effectiveness of meta-learning is due to its ability to learn a useful representation.
There is a line of works that analyzed the theoretical properties of meta-learning \citep{denevi2019learning,finn2019online,khodak2019adaptive,lee2019meta,bertinetto2018meta}.
We also note that there are analyses for other representation learning schemes \citep{arora2019theoretical,mcnamara2017risk,galanti2016theoretical,alquier2016regret,denevi2018incremental}.

 Linear bandits (stochastic linear bandits / linearly parameterized bandits / contextual linear bandits) have been studied in recent years \citep{auer2002using, dani2008stochastic, rusmevichientong2010linearly, abbasi2011improved, chu2011contextual, li2019nearly,li2019tight}. The studies are divided into two branches according to whether the action set is finite or infinite.
For the finite-action setting,  $\widetilde{\Theta}(\sqrt{dN})$ has been shown to be the near-optimal regret bound \citep{chu2011contextual, li2019nearly}, and for the infinite-action setting, $\widetilde{\Theta}(d \sqrt{N})$ regret bound has been shown to be near-optimal~\citep{dani2008stochastic, rusmevichientong2010linearly, li2019tight}.

\red{Some previous work studied the impact of low-rank structure in linear bandit. \citet{lale2019stochastic} studied a setting where the context vectors share a  low-rank structure. Specifically, in their setting, the context vectors consist of two parts, i.e. $\hat \vx = \vx + \bm \psi$, so that $\vx$ is from a hidden low-rank subspace and $\bm \psi$ is i.i.d. drawn from an isotropic distribution. 
	\citet{jun2019bilinear} and \citet{lu2020low} studied the bilinear bandits with low-rank structure. In their setting, the player chooses two actions $\vx, \vy$ and receives the stochastic reward with mean $\vx^\top \mTheta \vy$, where $\mTheta$ is an unknown low-rank bilinear form. The algorithms proposed in the aforementioned papers share some similarities with our \autoref{algo:etc} for our infinite-action setting, in that both used Davis-Kahan theorem to recover and exploit the low-rank structure.  }

Some previous work proposed multi-task bandits with different settings. \citet{deshmukh2017multi} proposed a setting under the contextual bandit framework. They assumed similarities among arms. \citet{bastani2019meta} studied a setting where the coefficients of the tasks were drawn from a  gaussian distribution fixed across tasks and proposed an algorithm based on Thompson sampling. \citet{soaremulti} proposed a setting where tasks were played one by one sequentially and the coefficients of the tasks were near in $\ell_2$ distance. In our setting, the tasks are played simultaneously and the coefficients share a common linear feature extractor. 
\section{Preliminaries}
\label{sec:pre}

\textbf{Notation.}
We use bold lowercases for vectors and bold uppercases for matrices.
For any positive integer $n$, we use $[n]$ to denote the set of integers $\{1,2,\ldots,n\}$. For any vector $\vx$, we use $\norm{\vx}$ to denote its $\ell_2$ norm. 
For a matrix $\mA$, we use $\norm{\mA}$ to denote the $2$-norm of $\mA$, $\norm{\mA}_{F}$ to denote the Frobenius norm, and $\norm{\mA}_{\max} = \max_{i, j} \abs{\mA_{ij}}$ to denote the max-norm. For two expressions $\alpha, \beta>0$, we denote $\alpha \lesssim \beta$ if there is a numerical constant $c > 0$ such that $\alpha \le c \beta$. We denote $\alpha \gtrsim \beta$ if $\beta \lesssim \alpha$. 

\textbf{Problem Setup.}
Let $d$ be the ambient dimension and $k (\le d)$ be the representation dimension.
In total, we have $T$ tasks and we play each task concurrently for $N$ rounds.
 Each task $t \in [T]$ has an unknown vector $\vtheta_t \in \sR^d$.  
At each round $n \in [N]$, the player chooses action $\va_{n, t} \in \gA_{n,t}$ for each task $t \in [T]$ where $\gA_{n,t}$ is the action set at round $n$ for the task $t$.

After the player commits to a batch of actions $\{\va_{n, t}\}_{t \in [T]}$, it receives a batch of rewards $\{r_{n, t}\}_{t \in [T]}$, where we assume $r_{n, t} = \langle \va_{n, t}, \vtheta_t \rangle + \varepsilon_{n, t}$.
Here we assume the noise $\varepsilon_{n, t}$ are independent $1$-sub-Gaussian random variables, which is a standard assumption in the literature. 

\red{We} use the total expected regret to measure the performance of our algorithm. When we have $N$ rounds and $T$ tasks, it is defined as
$
R^{N, T} = \sum_{n = 1}^N \sum_{t = 1}^T \max_{\va \in \gA_{n, t}} \langle \va, \vtheta_t \rangle - \langle \va_{n, t}, \vtheta_t \rangle.
$
When the action set is finite, we assume that all $\gA_{n, t}$ have the same size $K$, i.e. $\abs{\gA_{n, t}} \equiv K$. Furthermore, we write $\gA_{n, t} = \{\vx_{n, t, 1}, \ldots, \vx_{n, t, K}\}$. Besides, we interchangeably use the number $a_{n, t} \in [K]$ and the vector $\va_{n, t} = \vx_{n, t, a_{n, t}} \in \sR^d$ to refer to the same action.

\textbf{Assumptions.}
 Our main assumption is the existence of a common linear feature extractor.
 \begin{assum}[Common Feature Extractor]
 	\label{asmp:common}
There exists a linear feature extractor $\mB  \in \sR^{d \times k}$ and a set of linear coefficients $\{\vw_t\}_{t=1}^T$ such that the expected reward of the $t$-th task at the $n$-th round satisfies $\E[r_{t,n}] = \langle  \vw_t , \mB^\top \vx_{n,t,a_{n, t}}\rangle$.
 \end{assum}
 
For simplicity, we let $\mW = [\vw_1, \ldots, \vw_T]$. 
\autoref{asmp:common}  implies that  $\mTheta \triangleq [\vtheta_1, \ldots, \vtheta_T] = \mB \mW$.
Note this assumption is in a sense necessary to guarantee the effectiveness of representation learning because without it one cannot hope that representation learning helps.

In this paper, we mostly focus on the finite-action setting.
We put the following assumption on action sets.

\begin{assum} \label{assum:gauss} Marginally, for every $n \in [N], t \in [T], a \in [K]$, the contexts satisfy $\vx_{n, t, a} \sim \gN(\vzero, \Sigma_t)$ such that $\lambda_{\max}(\Sigma_t) \le O(1/d)$ and $\lambda_{\min}(\Sigma_t) \ge \Omega(1 / d)$. 
\end{assum}

\red{With} this assumption, we have an unknown covariance matrix $\Sigma_t$ for each task.
\red{At} each round, the actions of the $t$-th task are \red{sampled} from a Gaussian distribution with covariance $\Sigma_t$.
This is a prototypical setting for theoretical development on linear bandits with finite actions (See e.g., \cite{han2020sequential}).
At a population level, each one of the $K$ actions is equally good but being able to select different actions based on the realized contexts allows the player to gain more reward.

We will also study the infinite-action setting.
We first state our assumption about the action sets.
\begin{assum}[Ellipsoid Action Set] \label{assum:ellipsoid} We assume $\mathcal A_{n, t} = \gA_t = \{x^\top \mQ_t^{-1} x \le 1: x \in \mathbb R^d\}$ is an ellipsoid with $\lambda_{\min}(\mQ_t) \ge \lambda_0 = \Omega(1)$. 
\end{assum}
The first assumption states that each action set is an ellipsoid that covers all directions.
This is a standard assumption, e.g., see \cite{rusmevichientong2010linearly}.

In this setting, we will also need to put some additional assumptions on the underlying parameters $\mB$ and $\mW$.
\begin{assum}[Diverse Source Tasks] We assume that $\lambda_{\min}(\frac{1}{T}\mW \mW^\top) \ge \frac{\nu}{k}$, where $\nu = \Omega(1)$. \label{assum:diverse}
\end{assum}
This assumption roughly states that the underlying linear coefficients $\{\vw_t\}_{t = 1}^T$ equally spans all directions in $\sR^k$. 
This is a common assumption in representation learning literature that enables us to learn the linear feature extractor \citep{du2020few, tripuraneni2020provable}. 
For example, the assumption holds with high probability when $\vw_i$ is uniformly chosen from the sphere $\sS^{k-1}$.

\begin{assum} 
	\label{assum:w}
	We assume $ \lVert \vw_t \rVert \ge \omega = \Omega(1)$.
\end{assum}
This is a normalization assumption on the linear coefficients.

\section{Main Results for Finite-Action Setting}
\label{sec:upper_bound}

\begin{algorithm}[tbp]
	\caption{MLinGreedy: Multi-task Linear Bandit with Finite Actions}
	\label{algo:mlinucb}
	Let $M = \lceil \log_2 \log_2 N \rceil, \gG_0 = 0, \gG_M = N, \gG_m = N^{1 - 2^{-m}}$ for $1 \le m \le M - 1$, let $\hatvtheta_{0, t} \gets \vzero$\;
	\For{$m \gets 1, \ldots, M$}{
		\For{$n \gets \gG_{m - 1} + 1, \ldots, \gG_m$}{
			For each task $t \in [T]$: choose action $a_{n, t} = \argmax_{a \in [K]}  \vx_{n, t, a}^\top \hatvtheta_{m - 1, t}$\; \label{loc:chooseaction}
		}
		Compute $\hatmB \hatmW \gets \argmin\limits_{\mB \in \sR^{d \times k}, \mW \in \sR^{k \times T}} \sum_{n = \gG_{m-1} + 1}^{\gG_m} \sum_{t = 1}^T [ \vx_{n, t, a_{n, t}}^\top  \mB \vw_t  - r_{n, t}]^2$\; \label{loc:line5}
		For each task $t \in [T]$: let $\hatvtheta_{m, t} = \hatmB \hatvw_{t}$\; \label{loc:linetheta}
	}
\end{algorithm}

In this section focus on the finite-action setting.
The pseudo-code is listed in \autoref{algo:mlinucb}. 
Our algorithm uses a doubling schedule rule \citep{gao2019batched, simchi2019phase, han2020sequential, ruan2020linear}.
We only update our estimation of $\vtheta$ after an epoch is finished, and we only use samples collected within the epoch.
In \autoref{loc:line5}, we solve an empirical $\ell_2$-risk minimization problem on the data collected in the last epoch to estimate the feature extractor $\mB$ and linear predictors $\mat{W}$, similar to \citet{du2020few}.
Given estimated feature extractor $\hatmB$ and linear predictors $\hatmW$, we compute our estimated linear coefficients of task $t$ as $\hatvtheta_t \triangleq \hatmB\hatvw_t$ in \autoref{loc:linetheta}.
For choosing actions, for each task, we use a greedy rule, i.e., we choose the action that maximizes the inner product with our estimated $\vtheta$ (cf. \autoref{loc:chooseaction}). 

The following theorem gives an upper bound on the regret of \autoref{algo:mlinucb}.

\begin{thm}[Regret of \autoref{algo:mlinucb}] \label{thm:ub} 
Suppose $K, T \le \mathrm{poly}(d)$ and $N \ge d^2$. 	
Under \autoref{asmp:common} and \autoref{assum:gauss},	the expected regret of \autoref{algo:mlinucb} is upper bounded by
\begin{align*}
    \E[R^{N, T}]  =  \widetilde{O}(T \sqrt{kN} + \sqrt{dkNT}). 
\end{align*}
\vspace{-0.5cm}
\end{thm}

There are two terms in \autoref{thm:ub}, and we interpret them separately.
The first term $\widetilde{O}(T\sqrt{kN})$ represents the regret for playing $T$ independent linear bandits \emph{with dimension $k$} for $N$ rounds.
This is the regret we need to pay even if we know the optimal feature extractor, with which  we can reduce the original problem to playing $T$ independent linear bandits with dimension $k$ (recall $\vw_t$ are different for different tasks).
The second term $\widetilde{O}(\sqrt{dkNT})$ represents the price we need to pay to learn the feature extractor $\mB$. 
Notably, this term shows we are using data across all tasks to learn $\mB$ as this term scales with $\sqrt{NT}$.

Now comparing with the naive strategy that plays $T$ independent $d$-dimensional linear bandits with regret $\widetilde{O}(T\sqrt{dN})$, our upper bound is smaller as long as $T = \Omega(k)$.
Furthermore, when $T$ is large, our bound is significantly stronger than $\widetilde{O}(T\sqrt{dN})$, especially when $k \ll d$.
To our knowledge, this is the first formal theoretical result showing the advantage of representation learning for bandit problems.
We remark that requiring $T=\Omega(k)$ is necessary.
One needs at least $k$ tasks to recover the span of $\mW$, so only in this regime representation learning can help.

Our result also puts a technical requirement on the scaling $K, T \le \mathrm{poly}(d)$ and $N \ge d^2$.
These are conditions that are often required in linear bandits literature. 
The first condition ensures that $K$ and $T$ are not too large, so we need not characterize $\log(KT)$ factors in regret bound. If they are too large, e.g. $K, T \ge \Omega(e^d)$, then we would have $\log KT = O(d)$ and we could no longer omit the logarithmic factors in regret bounds. The second condition ensures one can at least learn the linear coefficients up to \red{a} constant error.
See more discussions in \citep[Section 2.5]{han2020sequential}.

While \autoref{algo:mlinucb} is a straightforward algorithm, the proof of \autoref{thm:ub} requires a combination of representation learning and linear bandit techniques. First, we prove the in-sample guarantee of representation learning, as done in \autoref{lem:rep}. Second, we exploit \autoref{assum:gauss}  to show that the learned parameters could extrapolate well on new contexts, as shown in \autoref{lem:rep2}. The regret analysis then follows naturally. 
We defer the proof of \autoref{thm:ub} to \autoref{app:ub}.

The following theorem shows that \autoref{algo:mlinucb} is minimax optimal up to logarithmic factors.
\begin{thm}[Lower Bound for Finite-Action Setting] \label{thm:lb} Let $\gA$ denote an algorithm and $\gI$ denote a finite-actioned multi-task linear bandit instance that satisfies \autoref{asmp:common} and \autoref{assum:gauss}. Then for any $N, T, d, k \in \mathbb{Z}^+$ with $k \le d$, $k \le T$,  we have 
	\begin{align}
	\inf_{\gA} \sup_{\gI} \E[R^{N,T}_{\gA, \gI}]  = \Omega\left(T \sqrt{kN}+ \sqrt{dkNT}\right).
	\vspace{-0.5cm}
	\end{align}
\end{thm}

\autoref{thm:lb} has the exactly same two terms as in \autoref{thm:ub}.
This confirms our intuition that the two prices to pay are real: 1) playing $T$ independent $k$-dimensional linear bandits and 2) learning the $d\times k$-dimensional feature extractor.
We defer the proof of \autoref{thm:lb} to \autoref{app:lb}.
At a high level, we separately prove the two terms in the lower bound. The first term is established by the straightforward observation that our multi-task linear bandit problem is at least as hard as solving $T$ independent $k$-dimensional linear bandits. The second term is established by the observation that multi-task linear bandit can be seen as solving $k$ independent $d$-dimensional linear bandits, each has $NT / k$ rounds. Note that the observation would directly imply the regret lower bound $k \sqrt{d (NT / k)} = \sqrt{dkNT}$, which is exactly the second term. 
To our knowledge, this lower bound is also the first one of its kind for multi-task sequential decision-making problems.
We believe our proof framework can be used in proving lower bounds for other related problems.

\section{Extension to Infinite-Action Setting}
\label{sec:infarm}
\begin{algorithm}[tb]	
	\caption{$\mathsf{E^2TC}$: Explore-Explore-Then-Commit } \label{algo:etc}
	\KwIn{$N$: total number of rounds , $N_1$: number of rounds for stage 1 , $N_2$: number of rounds for stage 2}
\textit{Stage 1: Estimating Linear Feature Extractor with Method-of-Moments}\;
		\For{$\forall t \in [T], n \in [N_1]$}{
		 Play $x_{n, t} \sim \mathrm{Unif}(\lambda_0 \cdot \mathbb S^{d-1})$ and receive reward $r_{n, t}$\;
		}
	 Compute  $\widehat{\mat{M}} \gets \frac{1}{N_1 T} \sum_{n = 1}^{N_1} \sum_{t = 1}^T r_{n, t}^2  x_{n, t}  x_{n, t}^\top$\; \label{loc:mom}
	 Let $\hatmB\mD\hatmB^\top \gets$ top-$k$ singular value decomposition of $\widehat{\mM}$. Denote $\hatvb_i$ the $i$-th column of $\hatmB$\;
	 
 \textit{Stage 2: Estimating Optimal Actions on Low-dimensional Space }\;
 
\For{$\forall t \in [T], i \in [k]$}{
  Play $v_{t,i} \triangleq \sqrt{\lambda_0} \hat{b}_i$ for $N_2 / k$ times and receive rewards $\{r_{n, t}\}_{n=N_1 + (i-1)N_2/k+1 }^{N_1 + iN_2/k}$\;
  }
\For{$\forall t \in [T]$}{
 Estimate $\hatvw_t \gets \argmin_{\vw \in \mathbb R^k} \frac{1}{2N_2}\sum_{n = N_1 + 1}^{N_1 + N_2} [\langle x_{n, t} , \hatmB \vw \rangle - r_{n, t}]^2$\;
 Let $\hatvtheta_t = \hatmB \hatvw_t$\;
}
 \textit{Stage 3: Committing to Near-optimal Actions}\;
\For{$\forall t \in [T], n=N_1+N_2+1,\ldots,N$}{
  Play $\va_{n, t} \gets \argmax_{\va \in \gA_t} \langle a, \hatvtheta_t \rangle$ and receive reward $r_{n, t}$\;
}
\end{algorithm}

In this section we present and analyze an algorithm for the infinite-action setting.
Pseudo-code is listed in \autoref{algo:etc}.
Our algorithm has three stages, and we explain each step below.

\textbf{Stage 1: Estimating Linear Feature Extractor with Method-of-Moments.}
The goal of the first stage is to estimate the linear feature extractor $\mB$.
Our main idea to view this problem as a low-rank estimation problem for which we use \red{a}  method-of-moments estimator.
In more detail, we first sample each $x_{n,t} \sim \Unif[\lambda_0 \cdot \sS^d]$ for $N_1$ times.
We can use this sampling scheme because the action set is an ellipsoid.
Note this sampling scheme has a sufficient coverage on all directions, which help us estimate $\mB$.
Next, we compute the empirical weighted covariance matrix $
\hatmM = \frac{1}{N_1 T} \sum_{n = 1}^{N_1} \sum_{t = 1}^T r_{n, t}^2  \vx_{n, t}  \vx_{n, t}^\top.
$
To proceed, we compute the singular value decomposition of $\hatmM$ and keep its top-$k$ column space as  our estimated linear feature extractor $\hatmB$, which is a sufficiently accurate estimator (cf. \autoref{thm:mom}).

\textbf{Stage 2: Estimating Optimal Actions on Low-Dimensional Space.}
In the second stage, we use our estimated linear feature extractor to refine our search space for the optimal actions.
Specifically, we denote $\hatmB=[\hatvb_1,\ldots,\hatvb_k]$ and for $i \in [k]$ and $t \in [T]$, we let $\vv_{t,i} =\sqrt{\lambda_{0}}\hatvb_i$.
Under \autoref{assum:ellipsoid}, we know $\vv_{t,i} \in \gA_t$ for all $i \in [k]$.
Therefore, we can choose $\vv_{t,i}$ to explore.
Technically, our choice of actions $\vv_{t, i}$ also guarantees a sufficient cover in the sense that $
\lambda_{\min}\left(\sum_{i = 1}^k \hatmB^\top \vv_{t, i} \vv_{t, i}^\top \hatmB\right) \ge \lambda_0.
$
In particular, this coverage is on a \emph{low-dimensional space} instead of the original ambient space.

The second stage has $N_2$ rounds and on the $t$-th task, we just play each $\vv_{t,i}$ for $N_2 / k$ rounds. After that, we use linear regression to estimate $\vw_t$ for each task. Given the estimation $\hatvw_t$, we can obtain an estimation to the true linear coefficient $\hatvtheta_t \triangleq \hatmB \hatvw_t$.

\textbf{Stage 3: Committing to Near-Optimal Actions.}
After the second stage, we have an estimation $\hatvtheta_t$ for each task.
For the remaining $(N-N_1 - N_2)$ rounds, we just commit to the optimal action indicated by our estimations. Specifically, we play the action $\va_{n,t} \gets \argmax_{\va \in \gA_t}\langle \va_t, \hatvtheta_t\rangle$ for round $n = N_1+N_2+1,\ldots,N$.

\begin{figure}[!t]
	\centering
	\resizebox{\textwidth}{!}{\input{exp2/fig_syn_d30.pgf}}
	\caption{Comparisons of \autoref{algo:mlinucb} with the naive algorithm for $d=30$ on synthetic data.\label{fig:sim}}
	\resizebox{\textwidth}{!}{\input{exp2/fig_syn_d20.pgf}}
	\caption{Comparisons of \autoref{algo:mlinucb} with the naive algorithm for $d = 20$ on synthetic data.\label{fig:sim2}}
    \vspace{-0.5cm}
\end{figure}
The following theorem characterizes the regret of our algorithm.
\begin{thm}[Regret of \autoref{algo:etc}] \label{thm:regret} If we choose $N_1 = c_1d^{1.5}k \sqrt{\frac{N}{T}}$ and $N_2 = c_2  k \sqrt{N}$ for some constants $c_1,c_2 > 0$.
	The regret of \autoref{algo:etc} is upper bounded by \[\E[R^{N, T}] =  \widetilde{O}(d^{1.5} k \sqrt{TN} + k T \sqrt N). 
		\vspace{-0.3cm}
	\]
\end{thm}
The first term represents the regret incurred by estimating the linear feature extractor.
Notably, the term scales with $\sqrt{TN}$, which means we utilize all $TN$ data here.
The first term scales with $d^{1.5}k$, which we conjecture is unavoidable at least by our algorithm.
The second term represents playing $T$ independent $k$-dimensional infinite-action linear bandits.

Notice that if one uses a standard algorithm, e.g. the PEGE algorithm  \citep{rusmevichientong2010linearly}, to play  $T$ tasks independently, one can achieve an $\widetilde{O}(dT\sqrt{N})$ regret.
Comparing with this bound, our bound's second term is always smaller and the first is smaller when $T = \widetilde{\Omega}(dk^2)$.
This demonstrates that more tasks indeed help us learn the representation and reduce the regret.

We complement our upper bound with a lower bound below.
This theorem suggests our second term is tight but there is still a gap in the first term.
We leave it as an open problem to design new algorithm to match the lower bound or to prove a stronger lower bound.
\begin{thm}[Lower Bound for Infinite-Action Setting] \label{thm:inflb} Let $\gA$ denote an algorithm and $\gI$ denote an infinite-action multi-task linear bandit instance that satisfies \autoref{asmp:common}, \autoref{assum:ellipsoid}, \autoref{assum:diverse}, \autoref{assum:w}. Then for any $N, T, d, k \in \mathbb{Z}^+$ with $k \le d$, $k \le T$,  we have 
	\begin{align}
	\inf_{\gA} \sup_{\gI} \E[R^{N,T}_{\gA, \gI}]  = \Omega\left(d\sqrt{kNT} + kT \sqrt{N}\right).
	\vspace{-0.5cm}
	\end{align}
\end{thm}

\section{Experiments}

\label{sec:exp}

\begin{figure}[!t]
	\centering
	\resizebox{\textwidth}{!}{\input{exp2/fig_mnist_T10.pgf}}
	\caption{Comparisons of \autoref{algo:mlinucb} with the naive algorithm for $T = 10$ on MNIST.\label{fig:mnistsim}}
	\resizebox{\textwidth}{!}{\input{exp2/fig_mnist_T45.pgf}}
	\caption{Comparisons of \autoref{algo:mlinucb} with the naive algorithm for $T=45$ on MNIST.\label{fig:mnistsim2}}
    \vspace{-0.5cm}
\end{figure}

In this section, we use synthetic data and MNIST data to illustrate our theoretical findings and demonstrate the effectiveness of our Algorithm for the finite-action setting.
We also have simulation studies for the infinite-action setting, which we defer to \autoref{sec:exp_infarm}.
\red{
The baseline is the naive algorithm which plays $T$ tasks independently, and for each task, thie algorithm uses linear regression to estimate $\vtheta_t$ and choose the action greedily according to the estimated $\vtheta_t$.
	}

\subsection{Synthetic Data}

\textbf{Setup.} The linear feature extractor  $\mB$ is uniformly drawn from the set of $d\times k$ matrices with orthonormal columns.\footnote{We uniformly (under the Haar measure) choose a random element from the orthogonal group $O(d)$ and uniformly choose $k$ of its columns to generate $\mB$.} Each linear coefficient $\vw_t$ is uniformly chosen from the $k$-dimensional sphere. The noises are i.i.d. Gaussian: $\varepsilon_{n, t, a} = \gN(0, 1)$ for every $n \in [N], t \in [T], a \in [K]$. 
We fix $K=5$ and $N=10000$ for all simulations on finite-action setting.
We vary $k$, $d$ and $T$ to compare \autoref{algo:mlinucb} and the naive algorithm.

\textbf{Results and Discussions.} We present the simulation results in \autoref{fig:sim} and \autoref{fig:sim2}. We emphasize that the $y$-axis in our figures corresponds to the regret per task, which is defined as $R^{N, T} / T$. We fix $K = 5, N = 10000$. 
These simulations verify our theoretical findings.
First, as the number of tasks increases, the advantage of our algorithm increases compared to the naive algorithm.
Secondly, we notice that as $k$ becomes larger (relative to $d$), the advantage of our algorithm becomes smaller.
This can be explained by our theorem that as $k$ increases, our algorithm pays more regret, whereas the naive algorithm's regret does not depend on $k$.

\subsection{Finite-Action Linear Bandits for MNIST}
\textbf{Setup.} We create a linear bandits problem on MNIST data~\citep{lecun2010mnist} to illustrate the effectiveness of our algorithm on real-world data. 
We fix $K = 2$ and create $T = \binom{10}{2}$ tasks and each task is parameterized by a pair $(i, j)$, where $0 \le i < j \le 9$. We use $\gD_i$ to denote the set of MNIST images with digit $i$. At each round $n \in [N]$, for each task $(i, j)$, we randomly choose one picture from $\gD_i$ and one from $\gD_j$, then we present those two pictures to the algorithm and assign the picture with larger digit with reward $1$ and the other with reward $0$. The algorithm is now required to select an image (action). 
We again compare our algorithm with the naive algorithm.

\textbf{Results and Discussions.} 
The experimental results are displayed in \autoref{fig:mnistsim} for $T=10$ (done by constructing tasks with  first five digits) and \autoref{fig:mnistsim2} for $T=45$.
We observe for both $T=10$ and $T=45$, our algorithm significantly outperforms the naive algorithm for all $k$. 
Interestingly, unlike our simulations, we find the advantage of our algorithm does not decrease as we increase $k$.
We believe the reason is the optimal predictor is not exactly linear, and we need to develop an agnostic theory to explain this phenomenon, which we leave as a future work.
\section{Conclusion}
\label{sec:conclusion}
We initiate the study on the benefits of representation learning in bandits. 
We proposed new algorithms and demonstrated that in the multi-task linear bandits, if all tasks share a common linear feature extractor, then  representation learning provably reduces the regret.
We demonstrated empirical results to corroborate our theory.
An interesting future direction is to generalize our results to general reward function classes~\citep{li2017provably, agrawal2019mnl}.
\red{
In the following, we discuss some future directions.}

\red{
\paragraph{Adversarial Contexts}
For the finite-action setting, we assumed the context are i.i.d. sampled from a Gaussian distribution. In the bandit literature, there is a large body on developing low-regret algorithms for the adversarial contexts setting.
We leave it as an open problem to develop a algorithm with an $\widetilde{O}(T\sqrt{kN} + \sqrt{dkNT})$ upper bound or show this bound is not possible in the adversarial contexts setting.
One central challenge for the upper bound is that existing analyses for multi-task representation learning requires i.i.d. inputs even in the supervised learning setting. Another challenge is how to develop a confidence interval for an unseen input in the multi-task linear bandits setting. This confidence interval should utilize the common feature extractor and is tighter than the standard confidence interval for linear bandits, e.g. LinUCB.
	}

\red{
	\paragraph{Robust Algorithm}
The current approach is tailored to the assumption that there exists a common feature extractor. One interesting direction is to develop a robust algorithm. For example, consider the scenario where whether there exists a common feature extractor is unknown. We want to develop an algorithm with regret bound as in this paper when the common feature extractor exists and gracefully degrades to the regret of $T$ independent linear bandits when the common feature extractor does not exist. 
	}

\red{
\paragraph{General Function Approximation}
In this paper, we focus on linear bandits.
In the bandits literature, sublinear regret guarantees have been proved for more general reward function classes beyond the linear one \citep{li2017provably, agrawal2019mnl}.
Similarly, in the supervised representation learning literature,  general representation function classes have also been studied \citep{maurer2016benefit,du2020few}.
An interesting future direction is to merge these two lines of research by developing  provably efficient algorithms for multi-task bandits problems where general function class is used for representation.
}

\subsubsection*{Acknowledgments}

The authors would like to thank the anonymous reviewers for their comments and suggestions on our paper. WH is supported by NSF, ONR, Simons Foundation, Schmidt Foundation, Amazon Research, DARPA and SRC. JDL acknowledges support of the ARO under MURI Award W911NF-11-1-0303, the Sloan Research Fellowship, and NSF CCF 2002272. 

\bibliographystyle{iclr2021_conference}

\newpage
\appendix
\section{Proof of \autoref{thm:ub}}
\label{app:ub}

\begin{lem}[General Hoeffding's inequality, \cite{vershynin2018high}, Theorem 2.6.2] \label{lem:genhoef} Let $X_1, \ldots, X_n$ be independent random variables such that $\E[X_i] = 0$ and $X_i$ is $\sigma_i$-sub-Gaussian. Then there exists a constant $c > 0$, such that for any $\delta > 0$, we have 
\begin{align}
\Pr[\abs{\sum_{i = 1}^n X_i} \ge c \sqrt{\sum_{i = 1}^n \sigma_i^2 \log(1/\delta)}] \le \delta.
\end{align}
\end{lem}

\begin{lem} \label{lem:rep} Let $T$ be the number of tasks and $N_0$ be the number of samples. For every $(n, t) \in [N_0] \times [T]$, let $\vx_{n, t} \in \sR^d$ be fixed vectors and let $y_{n, t} = \vx_{n, t}^\top \vtheta_t + \varepsilon_{n, t}$, where $\vtheta_t \in \sR^d$ is a vector and $\varepsilon_{n, t}$ is an independent 1-sub-Gaussian variable. Let 
\begin{align}
    \hatmB \hatmW = \argmin\limits_{\hatmB \in \sR^{d \times k}, \hatmW \in \sR^{k \times T}} \sum_{n = 1}^{N_0} \sum_{t = 1}^T [ \vx_{n, t}^\top  \hatmB \hatvw_t  - y_{n, t}]^2, \label{eq:rep-1000}
\end{align}
where $\hatmW = \mqty(\hatvw_1 & \cdots & \hatvw_T)$. Then with probability $1 - \delta$, we have 
\begin{align}
    \sum_{n = 1}^{N_0} \sum_{t = 1}^T [\vx_{n, t}^\top (\hatmB \hatvw_t - \mB \vw_t)]^2 \lesssim (dk + kT)\log(NTdk) + \log(1/\delta). \notag
\end{align}
\end{lem}

\begin{proof} By \eqref{eq:rep-1000}, we have 
\begin{align}
    \sum_{n = 1}^{N_0} \sum_{t = 1}^T [ \vx_{n, t}^\top \hatmB \hatvw_t - y_{n, t}]^2 \le 
    \sum_{n = 1}^{N_0} \sum_{t = 1}^T [ \vx_{n, t}^\top \mB \vw_t - y_{n, t}]^2. \notag
\end{align}
Since $y_{n, t} = \vx_{n, t}^\top \mB \vw_t + \varepsilon_{n, t}$, we have 
\begin{align}
    \sum_{n = 1}^{N_0} \sum_{t = 1}^T [ \vx_{n, t}^\top (\hatmB \hatvw_t - \mB \vw_t) + \varepsilon_{n, t}]^2 \le 
    \sum_{n = 1}^{N_0} \sum_{t = 1}^T \varepsilon_{n, t}^2, \notag
\end{align}
which implies 
\begin{align}
    \sum_{n = 1}^{N_0} \sum_{t = 1}^T [\vx_{n, t}^\top (\hatmB \hatvw_t - \mB \vw_t)]^2 \le \sum_{n = 1}^{N_0} \sum_{t = 1}^T 2 \varepsilon_{n, t}\vx_{n, t}^\top (\hatmB \hatvw_t - \mB \vw_t). \label{eq:rep-2000}
\end{align}

Next we bound the right-hand side of \eqref{eq:rep-2000} via a uniform concentration argument. We let 
\begin{align}
    \gB = \{\mB' \in \sR^{d \times k} : \norm{\mB'}_{\max} \le 1\}, \qquad \gW = \{\mW' \in \sR^{k \times T} : \norm{\mW'}_{\max} \le 1 \}. \notag
\end{align}
For any fixed matrices $\mB' \in \gB, \mW' \in \gW$, we write $\mW' = \mqty(\vw'_1 & \cdots & \vw'_T)$ and define
\begin{align}
    \eta_{n, t}(\mB', \mW')= 2 \varepsilon_{n, t}\vx_{n, t}^\top (\mB' \vw'_t - \mB \vw_t). \notag
\end{align}
Note that $\eta_{n, t}(\mB', \mW')$ is an independent sub-Gaussian variable with sub-Gaussian norm $2\vx_{n, t}^\top (\mB' \vw'_t - \mB \vw_t)$. By the general Hoeffding's inequality (\autoref{lem:genhoef}), with probability $1 - \delta$, we have 
\begin{align}
    & \qquad ~~ \sum_{n = 1}^{N_0} \sum_{t = 1}^T \eta_{n, t}(\mB', \mW') \lesssim \sqrt{ f(\mB', \mW')  \log(1/\delta)},  \label{eq:rep-3000} \\ 
    & \text{where} ~ f(\mB', \mW') = \sum_{n = 1}^N \sum_{t = 1}^T [ \vx_{n, t}^\top (\mB' \vw'_t - \mB \vw_t) ]^2. \notag
\end{align}
Next we apply the $\epsilon$-net. Let $\gB' = \gN(\gB, \norm{\cdot}_{\max}, \epsilon), \gW' = \gN(\gW, \norm{\cdot}_{\max}, \epsilon)$. Applying an union bound over $\gB' \times \gW'$ for \eqref{eq:rep-3000}, we have
\begin{align}
    \Pr[\forall (\mB', \mW') \in \gB' \times \gW': \sum_{n = 1}^{N_0} \sum_{t = 1}^T \eta_{n, t} \lesssim \sqrt{ f(\mB', \mW')  \log(1/\delta)}] \ge 1 - \delta \abs{\gB' \times \gW'}. \notag 
\end{align}
Since $f(\mB', \mW')$ is $(3NTdk)$-Lipschitz with respect to  $\norm{\cdot}_{\max}$, we have 
\begin{align}
\Pr[ \forall (\mB', \mW') \in \gB \times \gW:  \sum_{n = 1}^{N_0} \sum_{t = 1}^T \eta_{n, t} \lesssim \sqrt{ f(\mB', \mW')  \log(1/\delta)} + 24NTdk \epsilon] \ge 1 - \delta \abs{\gB' \times \gW'}. \notag
\end{align}
Note that $\abs{\gB'} = O((1/\epsilon)^{dk})$ and $\abs{\gW'} = O((1/\epsilon)^{kT})$. Let $\delta = \delta_0 / \abs{\gB' \times \gW'}$ and $\epsilon = (24NTdk)^{-1}$. We have  
\begin{align}
    \Pr[ \forall (\mB', \mW') \in \gB \times \gW:  \sum_{n = 1}^{N_0} \sum_{t = 1}^T \eta_{n, t} \lesssim \sqrt{ f(\mB', \mW') [(dk + kT)\log(NTdk) + \log (1/\delta_0)]}] \ge 1 - \delta_0. \label{eq:rep-5000}
\end{align}
Now we assume that the above event holds. Combining \eqref{eq:rep-2000} and \eqref{eq:rep-5000}, we have 
\begin{align}
    &\quad \Pr[ \sqrt{f(\hatmB, \hatmW)} \lesssim \sqrt{ (dk + kT) \log(NTdk) + \log(1/\delta_0)}] \notag \\
    &= \Pr[ f(\hatmB, \hatmW) \lesssim \sqrt{ f(\hatmB, \hatmW) [(dk + kT) \log(NTdk) + \log(1/\delta_0)]}] \notag \\ 
    &\ge 1 - \delta_0, \notag
\end{align}
which proves our lemma.
\end{proof}

\begin{lem} \label{lem:anti} With probability $1 - O((NT)^{-2})$, for every $m \in [M], t \in [T]$, we have 
\begin{align}
    \lambda_{\min}\left( \sum_{n = \gG_{m - 1} + 1}^{\gG_m} \vx_{n, t, a_{n, t}}\vx_{n, t, a_{n, t}}^\top \right)  \gtrsim \frac{\gG_{m} - \gG_{m-1}}{d}. \notag
\end{align}

\end{lem}

\begin{proof} The proof can be done by following the proof of Lemma 4 in \citet{han2020sequential}.
\end{proof}

\begin{lem}  \label{lem:rep2} For each epoch $m \in [M]$, with probability $1 - O((NT^{-2}))$, we have 
\begin{align}
    \norm{\hatmB \hatmW - \mB \mW}_F^2 \lesssim \frac{(dk + kT)\log(NTdk) + \log(1/\delta)}{  (\gG_m - \gG_{m - 1}) / d}, \notag
\end{align}
where $\hatmB, \hatmW$ are computed at \autoref{loc:line5} in \autoref{algo:mlinucb}.
\end{lem}

\begin{proof} Placing $N_0 = \gG_m - \gG_{m-1}$ in \autoref{lem:rep}, with probability $1 - O((NT)^{-2})$, we have  
\begin{align}
     \sum_{n = \gG_{m - 1}}^{\gG_m} \sum_{t = 1}^T [\vx_{n, t, a_{n, t}}^\top (\hatmB \hatvw_t - \mB \vw_t)]^2 \lesssim (dk + kT)\log(NTdk). \label{eq:rep3-1000}
\end{align}
By \autoref{lem:anti}, with probability $1 - O((NT)^{-2})$, we have 
\begin{align}
   &~~  \sum_{n = \gG_{m - 1}}^{\gG_m} \sum_{t = 1}^T [\vx_{n, t, a_{n, t}}^\top (\hatmB \hatvw_t - \mB \vw_t)]^2 \notag \\
    & = \sum_{t = 1}^T  (\hatmB \hatvw_t - \mB \vw_t)^\top \left(\sum_{n = \gG_{m - 1}}^{\gG_m} \vx_{n, t, a_{n, t}}\vx_{n, t, a_{n, t}}^\top \right) (\hatmB \hatvw_t - \mB \vw_t) \notag \\
    &\gtrsim \sum_{t = 1}^T  (\hatmB \hatvw_t - \mB \vw_t)^\top  \frac{\gG_m - \gG_{m - 1}}{d} (\hatmB \hatvw_t - \mB \vw_t) \notag \\ 
    &=   \frac{\gG_m - \gG_{m - 1}}{d} \norm{\hatmB \hatmW - \mB \mW}_F^2. \label{eq:rep3-2000}
\end{align}
We conclude by combining \eqref{eq:rep3-2000} with \eqref{eq:rep3-1000}.
\end{proof}

Let 
\begin{align*}
    R_m = \sum_{n = \gG_{m-1} + 1}^{\gG_m} \sum_{t = 1}^T \max_{a \in [K]} \langle \vx_{n, t, a}, \vtheta_t \rangle - \langle \vx_{n, t, a_{n, t}}, \vtheta_t \rangle
\end{align*}
be the regret incurred in the $m$-th epoch. We have the following lemma. 
\begin{lem} We have 
\begin{align*}
    \E[R_m] \lesssim  (\sqrt{NTdk} + T \sqrt{kN}) \sqrt{\log(NTdk) \log(NKT)}. 
\end{align*}
\end{lem}

\begin{proof} At round $n$ that belongs to epoch $m$, for task $t$, we have 
\begin{align*}
 \max_{a \in [K]} \vtheta_t^\top (\vx_{n, t, a} - \vx_{n, t, a_{n, t}}) &\le \max_{a \in [K]} \{\vtheta_{t}^\top (\vx_{n, t, a} - \vx_{n, t, a_{n, t}}) + \hatvtheta_{m - 1, t}^\top(\vx_{n, t, a_{n, t}} - \vx_{n, t, a})\} \\ 
 &= \max_{a \in [K]} (\vtheta_t - \hatvtheta_{m - 1, t})^\top \vx_{n, t, a} + (\hatvtheta_{m - 1, t} - \vtheta_t)^\top \vx_{n, t, a_{n, t}} \\ 
 &\le 2 \max_{a \in [K]} \abs{(\vtheta_t - \hatvtheta_{m - 1, t})^\top \vx_{n, t, a}}.
\end{align*}
By \autoref{assum:gauss} and a union bound over $K$ actions, $T$ tasks, and $(\gG_m - \gG_{m - 1})$ rounds, we have with probability $1 - (NKT)^{-2}$, for every $n \in (\gG_{m - 1}, \gG_m]$ and $t \in [T]$, 
\begin{align}
\max_{a \in [K]} \abs{(\vtheta_t - \hatvtheta_{m - 1, t})^\top \vx_{n, t, a}} \lesssim \norm{\vtheta_t - \hatvtheta_{m - 1, t}} \sqrt{\frac{\log(NKT)}{d}}. \label{eq:lem5-1000}
\end{align}
Using a union bound over \eqref{eq:lem5-1000} and \autoref{lem:rep2}, with probability $1 - O((NT)^{-2})$, the regret incurred in the $m$-th epoch is 
\begin{align}
    \E[R_m] &\lesssim \sum_{n=\gG_{m-1} + 1}^{\gG_m} \sum_{t = 1}^T \max_{a \in [K]} \vtheta_t^\top (\vx_{n, t, a} - \vx_{n, t, a_{n, t}})  \notag \\ 
    &\lesssim (\gG_m - \gG_{m-1}) \sum_{t = 1}^T  \norm{\vtheta_t - \hatvtheta_{m - 1, t}} \sqrt{\frac{\log(NKT)}{d}} \le \gG_m \sqrt{T} \norm{\hatmTheta - \mTheta}_F \sqrt{\frac{\log(NKT)}{d}} \label{eq:lem5-2000} \\
     &\le \gG_m \sqrt{T} \sqrt{\frac{ (dk + kT)\log(NTdk)}{(\gG_{m-1} - \gG_{m - 2}) /d}} \sqrt{\frac{\log(NKT)}{d}} \label{eq:lem5-3000} \\
     &\lesssim   \sqrt{NT (dk + kT) \log(NTdk) \log(NKT)} \label{eq:lem5-4000} \\
     &\lesssim (\sqrt{NTdk} + T \sqrt{kN}) \sqrt{\log(NTdk) \log(NKT)}, \notag \qedhere
\end{align}
where we denote $\hatmTheta = \mqty(\hatvtheta_{m - 1, 1} & \cdots & \hatvtheta_{m-1, T})$ in \eqref{eq:lem5-2000} and the second inequality in \eqref{eq:lem5-2000} uses Cauchy, \eqref{eq:lem5-3000} uses \autoref{lem:rep2}, \eqref{eq:lem5-4000} uses $ \gG_{m-1} \lesssim \gG_{m - 1} - \gG_{m-2}$ and $\gG_m / \gG_{m-1} \lesssim \sqrt{N}$. Since that  the above bound holds with probability $1 - O((NT)^{-2})$ and that the regret is bounded by $NT$, we prove the lemma.
\end{proof}

\begin{proof}[Proof of \autoref{thm:ub}] The regret is bounded by
\begin{align}
    \E[R^{N, T}] &= \sum_{m = 1}^M \E[R_m] \lesssim M(\sqrt{NTdk} + T \sqrt{kN}) \sqrt{\log(NTdk) \log(NKT)} \notag \\
    &= (\sqrt{NTdk} + T \sqrt{kN}) \sqrt{\log(NTdk) \log(NKT)} \log \log N. \qedhere \notag
\end{align}
\end{proof}

\section{Proof of \autoref{thm:lb}}

\label{app:lb}

In this appendix, we assume that $\Sigma_t = \mI$ in \autoref{assum:gauss} and that all noises are gaussian, i.e. $\varepsilon_{n, t} \sim \gN(0, 1)$ for all $n \in [N], t\in [T]$.

For each task $t \in [T]$, we  denote the regret incurred on task $t$ as 
\begin{align}
    R^{N, (t)} = \sum_{n = 1}^N \max_{a \in [K]} \langle \vx_{n, t, a}, \vtheta_t \rangle - \langle \vx_{n, t, a_{n, t}}, \vtheta_t \rangle. \notag
\end{align}

We divide \autoref{thm:lb} into the following two lemmas. 

\begin{lem} \label{lem:lb1} Under the setting of \autoref{thm:lb}, we have $\inf_\gA \sup_\gI \E[R^{N, T}_{\gA, \gI}]  \ge \Omega(T \sqrt{kN})$. 
\end{lem}

\begin{lem} \label{lem:lb2} Under the setting of \autoref{thm:lb}, we have $\inf_\gA \sup_\gI \E[R^{N, T}_{\gA, \gI}]  \ge \Omega(\sqrt{dkNT})$. 
\end{lem}

\begin{proof}[Proof of \autoref{thm:lb}] We combine  \autoref{lem:lb1} and \autoref{lem:lb2}. 
\end{proof}

Our proofs to the lemmas will be based on the lower bounds for the (single-task) linear bandit setting, which corresponds to the $T = 1$ case in our multi-task setting. For this single-task setting, we assume $k = d$ and $\mB = \mI_d$. We write the regret as $R^{N} = R^{N, 1}$ and call algorithms for the single-task setting as single-task algorithms.

\begin{lem}[\cite{han2020sequential}, Theorem 2] \label{lem:finlbsinglegaussian}Assume $N \ge d^2$ and $d \ge 2$. Let $\mathcal N(\mu, \Sigma)$ be the multivariate normal distribution with mean $\mu$ and covariance matrix $\Sigma$. There is a constant $C > 0$, such that for any single-task  algorithm $\gS$, we have  
\begin{align*}
    \sup_{\norm{\vw} \le 1} \E[R^N_{\gS, \gI}] \ge C \sqrt{dN},
\end{align*}
where $\gI$ is the instance with hidden linear coefficients $\vw$. 
\end{lem}

 Next we use it to prove \autoref{lem:lb1} and \autoref{lem:lb2}. The main idea to prove \autoref{lem:lb1} is to note that we can treat our setting as $T$ independent $k$-dimensional linear bandits. 

\begin{proof}[Proof of \autoref{lem:lb1}] Suppose there is an algorithm $\gA$ that achieves $\sup_{\gI} \E[R^{N, T}_{\gA, \gI}] \le C T \sqrt{kN}$. Then we have 
\begin{align*}
    \sup_{\norm{\vw_t} \le 1} \E\left[\sum_{t = 1}^T R^{N, (t)}_{\gA, \gI}\right]  \le  \sup_{\gI} \E[R^{N, T}_{\gA, \gI}]  \le C T \sqrt{kN}.
\end{align*}
Therefore, there exists $t \in [T]$ such that 
\begin{align*}
    \sup_{\gI} \E[R^{N, (t)}_{\gA, \gI}] \le \frac{1}{T} C T \sqrt{kN} = C \sqrt{kN},
\end{align*}
which contradicts \autoref{lem:finlbsinglegaussian}.
\end{proof}

\begin{proof}[Proof of \autoref{lem:lb2}] Suppose there is an algorithm $\gA$ that achieves $\sup_{\gI} \E[R^{N, T}_{\gA, \gI}] \le C \sqrt{dkNT}$. 
We complete the proof separately, based on whether $k \le \frac{d}{2}$ or not. Note that when $k > \frac{d}{2}$, the lower bound in \autoref{lem:lb1} becomes $\Omega(T  \sqrt{kN}) = \Omega(T \sqrt{dN})$. Since $T \ge k$, we have $ \sqrt{d k T N} \lesssim T  \sqrt{dN}$. Thus we conclude by \autoref{lem:lb1}. 

In the remaining, we assume $k \le \frac{d}{2}$. Without loss of generality, we assume that $d$ is even and that $2 k$ divides $T$. For $i = 1, \ldots, k$, we denote the regret of group $i$ as 
\begin{align}
    R^{N, ((i))}_{\gA, \gI}  = \sum_{t = (i - 1) T / k + 1}^{i T / k} R^{N, (t)}_{\gA, \gI}. \notag
\end{align} 
We consider instances such that tasks from the same group share the same hidden linear coefficients $\vtheta_t$. Since there are $k$ groups, we have 
\begin{align}
    \sup_{\norm{\vtheta_t} \le 1} \E\left[\sum_{i = 1}^k R^{N, ((i))}_{\gA, \gI}\right]  \le  \sup_{\gI} \E[R^{N, T}_{\gA, \gI}]  \le C \sqrt{dkNT}.
\end{align}
Therefore, there exists a group $i \in [k]$ such that 
\begin{align}
    \sup_{\gI_i} \E[R^{N, ((i))}_{\gA, \gI}] \le \frac{1}{k}C \sqrt{dkNT} = C \sqrt{\frac{d N T}{k}}, \label{eq:lb2-3000}
\end{align}
which means that the regret incurred in group $i$ is less than $C \sqrt{d NT/k}$. Since the tasks in group $i$ share the same hidden linear coefficients, they could be regarded as one large linear bandit problem. Since there are $T / k$ tasks in group $i$, the large linear bandit is played for $N \cdot T / k$ rounds. By \autoref{lem:finlbsinglegaussian}, the algorithm $\gA$ must have incurred $C \sqrt{d NT / k}$ regret on group $i$, which contradicts with  \eqref{eq:lb2-3000}.
\end{proof}

\section{Method-of-Moments Estimator under Bandit Setting}

\label{app:mom}

The following theorem shows the guarantee of the method-of-moments (MoM) estimator we used to find the linear feature extractor $\mB$. 
In this appendix, for a matrix $\mB$ with orthogonal columns, we write $\mB_\perp$ to denote its orthogonal complement matrix (a matrix whose columns are the orthogonal complement of those of $\mB$). 
 \begin{thm}[MoM Estimator] \label{thm:mom}
 Assume $N_1 T \gtrsim \mathrm{polylog}(N_1, T) \cdot \frac{d^{1.5} k}{\lambda_0 \nu}$. We have with probability at least $1 - (N_1 T)^{-100}$, 
 \begin{align}
     \norm{\hatmB_\perp^\top \mB} \lesssim \widetilde{O}(\frac{d^{1.5 }k}{\lambda_0 \nu \sqrt{N_1 T}}). \label{eqn:bpertb}
 \end{align} 
 \end{thm}
 The theorem guarantees that our estimated $\hatmB$ is close to the underlying $\mB$ in the operator norm so long as the values $N_1$ and $T$ are  sufficiently large. We add a remark that our theorem is similar to Theorem 3 in \cite{tripuraneni2020provable}. The key differences are:
(i) we use a uniform distribution to find the feature extractor, while they assumed the input distribution is standard $d$-dimensional Gaussian; (ii) the SNR (signal-to-noise ratio) in our linear bandit setting is worse than that in their supervised learning setting, and thus we get an extra $d$ factor in our theorem. 

In the sequel, we prove the theorem. 

\begin{lem}[Hoeffding] Let $\varepsilon_1, \ldots, \varepsilon_N$ be i.i.d. $1$-sub-Gaussian random variables. We have 
\begin{align*}
    \Pr[ \abs{\frac{1}{N}\sum_{i = 1}^N (\varepsilon_i -\E{\varepsilon_i}) } \ge t] \le 2 e^{-\frac{t^2}{2N}}.
\end{align*}

\end{lem}

\begin{lem}[Matrix Bernstein's inequality, \citet{vershynin2018high},  Theorem 5.4.1] \label{lem:mbi} Let $\mX_1, \ldots, \mX_m \in \sR^{d \times d}$ be independent, mean zero, symmetric random matrices that $\norm{\mX_i} \le M$ almost surely for all $i \in [m]$. 
Let $\sigma^2 = \norm{\sum_{i = 1}^m \E \mX_i^2}$. We have 
\begin{align*}
    \Pr[\norm{ \sum_{i = 1}^m \mX_i } \ge \delta] \le 2 d \exp(- \frac{\delta^2 / 2}{\sigma^2 + M \delta / 3}).
\end{align*}
Equivalently, with probability at least $1 - \delta$, we have 
\begin{align*}
    \norm{\sum_{i = 1}^m \mX_i} \lesssim \sqrt{\sigma^2 \log(d / \delta)} + M \log(d / \delta).
\end{align*}
\end{lem}

\begin{lem}[Moments of Uniform Distribution on Sphere] Let $\vx \sim \mathrm{Unif}(\mathbb S^{d-1})$ be a uniformly chosen unit vector. We have $\E{x_1^6} = \frac{15}{d (d + 2) (d + 4)}$, $\E{x_1^4} = \frac{3}{d(d + 2)}$, $\E{x_1^2} = \frac{1}{d}$. Moreover, we have $\E{x_1^4 x_2^2} = \frac{3}{d(d+2)(d + 4)}$.
\end{lem}

\begin{proof} We need to recall the fact that when $\vx\sim \Unif(\sS^{d-1})$, its coordinate $x_i$ follows the Beta distribution: $\frac{x_1 + 1}{2} \sim \mathrm{Beta}(\frac{d-1}{2}, \frac{d-1}{2})$. Then we prove the lemma by noting the moments of the Beta distribution \citep{fang2018symmetric}.
\end{proof}

\begin{coro}[Uniform Distribution on Sphere] Let $x \sim \mathrm{Unif}(\sS^{d-1})$ be a uniformly chosen unit vector. We have the following statements. 
\begin{enumerate}[label=(\alph*)]
\item $\E{\langle \vx, \vtheta \rangle^2 \vx \vx^\top} = \frac{2 \vtheta \vtheta^\top +  \mI}{d (d + 2)}$.
    \item $\E{\langle \vx, \vtheta \rangle^4 \vx \vx^\top} = \frac{12 \vtheta \vtheta^\top + 3 \mI}{d (d + 2) (d + 4)}$.
\end{enumerate}
\end{coro}

\begin{proof} Let $\ve_1 = (1, 0, \ldots, 0) \in \mathbb R^d$ be the unit vector. For (a), note that 
    \begin{align*}
    (\E{\langle \vx, \ve_1 \rangle^2 \vx \vx^\top})_{ij} = \E x_1^2 x_i x_j = \begin{cases}
    0, & i, j \ne 1 \text{ and }i \ne j, \\
    0, & i = 1 \ne j \text{ or } j = 1 \ne i, \\
    \frac{1}{d(d+2)},  & i = j \ne 1, \\
    \frac{3}{d (d + 2)},  & i = j = 1.
    \end{cases}
    \end{align*}
    Therefore, we have $\E[\langle \vx, \ve_1 \rangle^2 \vx \vx^\top] = \frac{1}{d (d + 2)}(2 \ve_1 \ve_1^\top +  \mI)$. By the isotropy (rotation invariance) of uniform distribution, we have
\begin{align*}
    \E{\langle \vx, \vtheta \rangle^2 \vx \vx^\top} = \frac{1}{d (d + 2)}(2 \vtheta \vtheta^\top +  \mI).
\end{align*}
For (b), note that  
    \begin{align*}
    (\E{\langle \vx, \ve_1 \rangle^4 \vx \vx^\top})_{ij} = \E x_1^4 x_i x_j = \begin{cases}
    0, & i, j \ne 1 \text{ and }i \ne j, \\
    0, & i = 1 \ne j \text{ or } j = 1 \ne i, \\
    \frac{3}{d(d+2)(d + 4)},  & i = j \ne 1, \\
    \frac{15}{d (d + 2) (d + 4)},  & i = j = 1.
    \end{cases}
    \end{align*}
    Therefore, we have $\E{\langle \vx, \ve_1 \rangle^4 \vx \vx^\top} = \frac{1}{d (d + 2) (d + 4)}(12 \ve_1 \ve_1^\top + 3 \mI)$. By the isotropy, we have  
    \[
    \E{\langle \vx, \vtheta \rangle^4 \vx \vx^\top} = \frac{1}{d (d + 2) (d + 4)}(12 \vtheta \vtheta^\top + 3 \mI). \qedhere
    \]

\end{proof}

Let $\mA_{n, t} = r_{n, t}^2  x_{n, t}  x_{n, t}^\top, \mM = \frac{1}{N_1 T} \sum_{n = 1}^{N_1} \sum_{t = 1}^T \mA_{n, t}$. We decompose $\mM$ into three terms $\mM = \mM_1 + \mM_2 + \mM_3$, where
\begin{align*}
\mM_1 &= \frac{1}{N_1 T} \sum_{n = 1}^{N_1} \sum_{t = 1}^T \langle \vx_{n, t}, \vtheta_t\rangle^2  \vx_{n, t} \vx_{n, t}^\top, \\
\mM_2 &= \frac{1}{N_1 T} \sum_{n = 1}^{N_1} \sum_{t = 1}^T 2 \varepsilon_{n, t} \langle \vx_{n, t}, \vtheta_t\rangle \vx_{n, t} \vx_{n, t}^\top, \\
\mM_3 &= \frac{1}{N_1 T} \sum_{n = 1}^{N_1} \sum_{t = 1}^T \varepsilon_{n, t}^2 \vx_{n, t} \vx_{n, t}^\top.
\end{align*}
Let $\mA_{1nt} = \frac{1}{N_1 T}\langle \vx_{n, t}, \vtheta_t\rangle^2  \vx_{n, t} \vx_{n, t}^\top$.  Next we analyze each error $\norm{\mM_i - \E{\mM_i}}$.

\begin{lem} \label{lem:m1} With probability at least $1 - \frac{1}{N_1^3 T^3}$, we have
\begin{align*}
    \norm{\mM_1 - \E{\mM_1}} \lesssim \sqrt{\frac{\log(d N_1 T)}{d^3 N_1 T}} + \frac{\log(d N_1 T)}{N_1 T}.
\end{align*}
\end{lem}

\begin{proof} We have
\begin{align*}
\E{\mA_{1nt}^2} = \frac{1}{N_1^2 T^2}\E{\langle \vx_{n, t}, \vtheta_t \rangle^4 \vx_{n, t} \vx_{n, t}^\top} 
= \frac{12 \vtheta_t \vtheta_t^\top + 3 \mI}{d (d + 2) (d + 4) N_1^2 T^2}.
\end{align*}
Using \autoref{lem:mbi} with $m = N_1 T, M = \frac{1}{N_1 T}, \sigma^2 \lesssim m \cdot \frac{1}{d^3 N_1^2 T^2} = \frac{1}{d^3 N_1 T}$, we have with probability at least $1 - \frac{1}{N_1^3 T^3}$, 
\[
    \norm{\mM_1 - \E{\mM_1}} \lesssim \sqrt{\frac{\log(d N_1 T)}{d^3 N_1 T}} + \frac{\log(d N_1 T)}{N_1 T}. \qedhere
\]
\end{proof}

Next we analyze the errors of $\mM_2$ and $\mM_3$. Since these errors contain the unbounded sub-Gaussian terms $\varepsilon_{n, t}$, we need to cut their tails before applying the matrix Bernstein inequality. Define $\varepsilon'_{n, t} = \varepsilon_{n, t} \ind\{ \lvert \varepsilon_{n, t} \rvert \le R\}$. We have 
\begin{align*}
    \left| \E{\varepsilon_{n, t}} - \E{\varepsilon'_{n, t}} \right| &\le \E{ \lvert \varepsilon_{n, t} \rvert \ind\{\lvert \varepsilon_{n, t}\rvert > R\}} \\
    &= R \cdot \Pr[\lvert \varepsilon_{n, t} \rvert > R] + \int_{R}^{+\infty} \Pr[\lvert \varepsilon_{n, t} \rvert > x] \dd{x} \\
    &\le 2Re^{-\frac{R^2}{2}} + \int_R^{+\infty} 2 e^{-\frac{x^2}{2}} \dd{x} \\
    &\le 2(R + \frac{1}{R}) e^{-\frac{R^2}{2}}.
\end{align*}
Let $\varepsilon''_{n, t} =  \varepsilon_{n, t}^2 \ind\{ \lvert \varepsilon_{n, t} \rvert \le R\}$. We have 
\begin{align*}
    \left| \E{\varepsilon_{n, t}^2} - \E{\varepsilon''_{n, t}} \right| &\le \E{ \varepsilon_{n, t}^2 \ind \{\varepsilon_{n, t}^2 > R^2\}} \\
    &= R^2 \cdot \Pr[ \varepsilon_{n, t}^2  > R^2] + \int_{R^2}^{+\infty} \Pr( \varepsilon_{n, t}^2 > x) \dd{d}x \\
    &\le 2 R^2 e^{-\frac{R^2}{2}} + \int_{R^2}^{+\infty} 2 e^{-\frac{x}{2}}\dd{d}x \\
    &\le (2 R^2 + 4) e^{- \frac{R^2}{2}}.
\end{align*}

Define 
\begin{align*}
\mM'_2 &= \frac{1}{N_1 T} \sum_{n = 1}^{N_1} \sum_{t = 1}^T 2 \varepsilon'_{n, t} \langle \vx_{n, t}, \vtheta_t\rangle \vx_{n, t} \vx_{n, t}^\top, \\
\mM''_3 &= \frac{1}{N_1 T} \sum_{n = 1}^{N_1} \sum_{t = 1}^T \varepsilon''_{n, t} \vx_{n, t} \vx_{n, t}^\top.
\end{align*}

\begin{lem} \label{lem:m2} With probability at least $1 - (N_1 T)^{-3}$, we have 
$$\norm{ \mM'_2 - \E{\mM'_2} } \lesssim \sqrt{\frac{\log(d N_1 T)}{d^2 N_1 T}} + \frac{R \log(d N_1 T)}{N_1 T}.$$
\end{lem}

\begin{proof} Let $\mA_{2nt} = \frac{1}{N_1 T}2\varepsilon'_{n, t} \langle \vx_{n, t}, \vtheta_t\rangle \vx_{n, t} \vx_{n, t}^\top$. We find that 
\begin{align*}
    \E{\mA_{2nt}^2} = \frac{4\E{(\varepsilon'_{n,t})^2}}{N_1^2 T^2} \E{\langle \vx_{n, t}, \vtheta_t\rangle^2 \vx_{n, t} \vx_{n, t}^\top} = \frac{4\E{ (\varepsilon'_{n,t})^2}}{N_1^2 T^2} \frac{2\vtheta_t \vtheta_t^\top + \mI}{d(d+2)}.
\end{align*}
We conclude by using \autoref{lem:mbi} with $m = N_1 T, M = \frac{R}{N_1 T}, \sigma^2 \lesssim m \cdot \frac{1}{d^2 N_1^2 T^2} = \frac{1}{d^2 N_1 T}$.
\end{proof}

\begin{lem} \label{lem:m3} With probability at least $1 - (N_1 T)^{-3}$, we have 
$$\lVert \mM_3'' - \E{\mM_3''} \rVert \lesssim \sqrt{\frac{\log(d N_1 T)}{d N_1 T}} + \frac{R^2 \log(d N_1 T)}{N_1 T}.$$
\end{lem}

\begin{proof} Let $\mA_{3nt} = \frac{1}{N_1 T}\varepsilon''_{n, t} \vx_{n, t} \vx_{n, t}^\top$. We find that 
\begin{align*}
    \E{\mA_{3nt}^2} = \frac{\E{(\varepsilon''_{n,t})^2}}{N_1^2 T^2} \E \vx_{n, t} x_{n, t}^\top = \frac{\E{ (\varepsilon''_{n,t})^2}}{N_1^2 T^2} \frac{\mI}{d}.
\end{align*}
We conclude by using \autoref{lem:mbi} with $m = N_1 T, M = \frac{R^2}{N_1 T}, \sigma^2 \lesssim m \cdot \frac{1}{d N_1^2 T^2} = \frac{1}{d N_1 T}$.
\end{proof}

\begin{lem} With probability at least $1 - (N_1 T)^{-2}$, we have 
$$\norm{ \mM - \E{\mM} } \lesssim \sqrt{\frac{\log(d N_1 T)}{d N_1 T}} + \frac{ \log^2(d N_1 T)}{N_1 T}.$$
\end{lem}

\begin{proof} Let $R = \sqrt{8 \log(N_1 T)}$. With probability at least $1 - (N_1 T)^{-3}$, we have $\abs{ \varepsilon_{n, t}} \le R$ for every $n \in [N_1]$ and $t \in [T]$. Note that in this case, we have $\mM_2 = \mM'_2$ and $\mM_3 = \mM''_3$. 
Using a union bound over \autoref{lem:m1}, \autoref{lem:m2}, and \autoref{lem:m3}, we have with probability at least $1 - (N_1 T)^{-2}$, 
\begin{align*}
    \lVert \mM - \E{ \mM} \rVert &\le \lVert \mM_1 - \E{ \mM_1} \rVert + \lVert \mM_2 - \E{ \mM_2} \rVert + \lVert \mM_3 - \E{ \mM_3} \rVert \\
    & = \lVert \mM_1 - \E{\mM_1} \rvert +
    \lVert \mM'_2 - \E{ \mM_2} \rVert + \lVert \mM''_3 - \E{ \mM_3} \rVert \\
     &\le \lVert \mM_1 - \E{\mM_1} \rVert 
+ \Vert \mM'_2 - \E{\mM'_2} \rVert
+ \lVert \E{\mM'_2} - \E{\mM_2} \rVert \\ 
 &\quad + \lVert \mM''_3 - \E{\mM''_3 }\rVert + \lVert \E{ \mM''_3} - \E{ \mM_3} \rVert   \\
    &\lesssim \sqrt{\frac{\log(d N_1 T)}{d N_1 T}} + \frac{ \log^2(d N_1 T)}{N_1 T}. 
    \qedhere
\end{align*}
\end{proof}

 \begin{proof}[Proof of \autoref{thm:mom}] We note that $\sigma_{k + 1}(\mM) - \sigma_{k + 1}(\E{ \mM} ) \le \lVert \mat E \rVert$. Under \autoref{assum:ellipsoid}, we have 
 \begin{align*}
     \E{ \mM_1} = \frac{2 \lambda_0}{d (d + 2) T}\mat \vtheta \mat \vtheta^\top + c_1  \mat I, \quad \E{ \mM_2} = 0, \quad \E{ \mM_3} = c_3 \mat I,
 \end{align*} where $c_1, c_3 \in \mathbb R$ are constants. Since $$\sigma_k(\frac{1}{T}\mat \vtheta \mat \vtheta^\top) - \sigma_{k + 1}(\frac{1}{T}\mat \vtheta \mat \vtheta^\top) = \sigma_k(\frac{1}{T}\mat W \mat W^\top) = \frac{\nu}{k},$$
 we have 
 \begin{align*}
     \sigma_k(\E{\mM}) - \sigma_{k + 1}(\E{\mM}) \asymp \frac{\lambda_0 \nu}{d^2 k}.
 \end{align*}
 
 Assume $N_1 T \gtrsim \mathrm{polylog}(N_1, T) \cdot dk$ so that $\lVert \mat E \rVert \le \frac{\lambda_0 \nu}{d^2 k}$. Together with Davis-Kahan sin $\vtheta$ theorem, we have with probability at least $1 - (N_1 T)^{-100}$, 
 \begin{align}
     \lVert \hatmB_\perp^\top \mB \rVert &\lesssim \frac{\lVert \hatmB_\perp^\top \mat E \mB \rVert}{\sigma_k(\E \mM) - \sigma_{k + 1}(\E \mM) - \lVert \mat E \rVert} \label{eq:sintheta} \\ \notag
     &\le \frac{\lVert \mat E \rVert}{\sigma_k(\E \mM) - \sigma_{k + 1}(\E \mM) - \lVert \mat E \rVert} \lesssim \frac{\lVert \mat E \rVert}{\lambda_0 \nu / d^2 k} \\ \notag
     &\lesssim \frac{d^2 k}{\lambda_0 \nu} (\sqrt{\frac{\log(d N_1 T)}{d N_1 T}} + \frac{ \log^2(d N_1 T)}{N_1 T}) \lesssim \frac{d^{1.5}k}{\sqrt{N_1 T}} \cdot \mathrm{polylog}(d, N, T), 
     \end{align}
     where \eqref{eq:sintheta} uses the Davis-Kahan sin $\vtheta$ theorem \citep[][Section VII.3]{bhatia2013matrix}.
 \end{proof}

\section{Proof of \autoref{thm:regret}}

\label{app:regret}

Our proof is similar to the proof of Theorem 3.1 of \citet{rusmevichientong2010linearly}.

\begin{lem}[\citet{rusmevichientong2010linearly}, Lemma 3.5] \label{lem:norm} For two vectors $\vu, \vv \in \sR^d$, we have 
\begin{align*}
    \left\lVert \frac{\vu}{\lVert \vu \rVert} - \frac{\vv}{\lVert \vv \rVert} \right\rVert \le \frac{2\lVert \vu - \vv \rVert}{\max\{\lVert \vu \rVert, \lVert \vv \rVert\}}.
\end{align*}
\end{lem}

\begin{lem} \label{lem:commit} Let $x_t = \argmax_{x \in \mathcal A_t} \langle x, \hat \vtheta_t \rangle$. We have 
	\begin{align*}
	\max_{x \in \mathcal A_t} \langle x, \vtheta_t \rangle - \langle x_t, \vtheta_t \rangle \le J \frac{\lVert \vtheta_t - \hat \vtheta_t \rVert^2}{\lVert \vtheta_t \rVert}.
	\end{align*}
\end{lem}

\begin{proof} For $\vtheta \in \mathbb R^d$, we define $f_t(\vtheta) = \max_{\va \in \gA_t} \langle \va, \vtheta \rangle$. Let $\vx_t^* = \argmax_{\vx \in \mathcal A_t} \langle \vx, \vtheta_t \rangle$. Then we have 
\begin{align}
    \max_{\vx \in \gA_t} \langle \vx, \vtheta_t \rangle - \langle \vx_t, \vtheta_t \rangle &= \langle \vx_t^* - \vx_t, \vtheta_t \rangle = \langle \vx_t^*, \vtheta_t - \hatvtheta_t \rangle + \langle \vx_t^* - \vx_t, \hatvtheta_t \rangle + \langle \vx_t, \hatvtheta_t - \vtheta_t \rangle \notag \\
    &\le \langle \vx_t^*, \vtheta_t - \hatvtheta_t \rangle + \langle \vx_t, \hatvtheta_t - \vtheta_t \rangle = \langle \vx_t^* - \vx_t, \vtheta_t - \hatvtheta_t \rangle \notag \\
    &= \langle f_t(\vtheta_t) - f_t(\hatvtheta_t), \vtheta_t - \hatvtheta_t \rangle = \langle f_t(\frac{\vtheta_t}{\lVert \vtheta_t \rVert}) - f_t(\frac{\hatvtheta_t}{\lVert \hatvtheta_t \rVert}), \vtheta_t - \hatvtheta_t \rangle \notag \\ 
    &\le \lVert  f_t(\frac{\vtheta_t}{\lVert \vtheta_t \rVert}) - f_t(\frac{\hatvtheta_t}{\lVert \hatvtheta_t \rVert})\rVert \cdot \lVert \vtheta_t - \hatvtheta_t \rVert \label{eq:commit-1000} \le J \lVert \frac{\vtheta_t}{\lVert \vtheta_t \rVert} - \frac{\hat \vtheta_t}{\lVert \hat \vtheta_t \rVert} \rVert \cdot \lVert \vtheta_t - \hat \vtheta_t \rVert \\
    &\le 2J \frac{\lVert \vtheta_t - \hat \vtheta_t \rVert^2}{\lVert \vtheta_t \rVert}, \label{eq:commit-2000}
\end{align}
where the first inequality in \eqref{eq:commit-1000} uses Cauchy and \eqref{eq:commit-2000} uses \autoref{lem:norm}.
\end{proof}

\begin{lem} \label{lem:errdecomp} For each task $t \in [T]$, we have $\E \lVert \hatvtheta_t - \vtheta_t \rVert^2 \lesssim \frac{k^2}{\lambda_1^2 N_2} + \lVert \hatmB_\perp^\top \mB \rVert^2$.
\end{lem}
\begin{proof} We define $\vtheta_t' = \hatmB \hatmB^\top \vtheta_t$. Note that $\vtheta_t = \vtheta_t' + \hatmB_\perp \hatmB_\perp^\top$. We have
\begin{align*}
    \hatvtheta_t - \vtheta_t &= \hatmB \hatvw_t - (\hatmB \hatmB^\top \vtheta_t + \hatmB_\perp \hatmB_\perp^\top) \vtheta_t \\
    &= \hatmB (\hatvw_t - \hatmB^\top \vtheta_t) - \hatmB_\perp \hatmB_\perp^\top \mB \vw_t.
\end{align*}
Note that $\hatmB$ is perpendicular to $\hatmB_\perp$ and that $\lVert \hatmB \rVert = \lVert \hatmB_\perp \rVert = 1$.  We have 
\begin{align*}
    \lVert \hatvtheta_t - \vtheta_t \rVert^2 = \lVert \hatmB (\hatvw_t - \hatmB^\top \vtheta_t) \rVert^2 + \lVert \hatmB_\perp \hatmB_\perp^\top \mB w_t \rVert^2 \le \lVert \hat w_t - \hatmB^\top \vtheta_t \rVert^2 + \lVert \hatmB_\perp^\top \mB \rVert^2. 
\end{align*}
Let $v_i = \sqrt{\lambda_0} \hat{b}_i$. The OLS estimator is given by 
\begin{align*}
    \hat w_t &= \left(\sum_{n = N_1 + 1}^{N_1 + N_2 + 1} \hatmB^\top \vv_i \vv_i^\top \hatmB \right)^{-1} \sum_{n = N_1 + 1}^{N_1 + N_2} \hatmB^\top \vx_{n, t} r_{n, t} \\
    &= \left(\sum_{n = N_1 + 1}^{N_1 + N_2 + 1} \hatmB^\top \vx_{n, t} \vx_{n, t}^\top \hatmB \right)^{-1} \sum_{n = N_1 + 1}^{N_1 + N_2} \hatmB^\top \vx_{n, t} (\vx_{n, t}^\top \hatmB \vw'_t + \varepsilon_{n, t}) \\
    &= w'_t + \left(\sum_{n = N_1 + 1}^{N_1 + N_2 + 1} \hatmB^\top \vx_{n, t} \vx_{n, t}^\top \hatmB \right)^{-1} \sum_{n = N_1 + 1}^{N_1 + N_2} \hatmB^\top \vx_{n, t} \varepsilon_{n, t}.
\end{align*}
Write $\mA = \sum_{n = N_1 + 1}^{N_1 + N_2 + 1} \hatmB^\top \vx_{n, t} \vx_{n, t}^\top \hatmB$.
\begin{align*}
    \E\lVert \hatvw_t - \vw'_t \rVert^2 &= \sum_{n = N_1 + 1}^{N_1 + N_2} \vx^\top_{n, t} \hatmB \mA^{-2} \hatmB^\top \vx_{n, t} \E\varepsilon_{n, t}^2 \\
    &\le \sum_{n = N_1 + 1}^{N_1 + N_2} \vx^\top_{n, t} \hatmB \mA^{-2} \hatmB^\top \vx_{n, t}  \le N_2 \lVert \mA^{-2} \rVert  \le N_2 (\frac{k}{\lambda_0 N_2})^2 = \frac{k^2}{\lambda_0^2 N_2}.
\end{align*}

Putting together, we have 
\[
    \E\lVert \hatvtheta_t - \vtheta_t \rVert^2 \le \frac{k^2}{\lambda_0^2 N_2} + \lVert \hatmB_\perp^\top \mB \rVert^2. \qedhere
\]
\end{proof}

\begin{proof}[Proof of \autoref{thm:regret}] Note that $J = O(1)$ under our assumptions, as indicated by \citet{rusmevichientong2010linearly}. So we have 
\begin{align*}
    \E[R^{N,T}] &\le T N_1 + T N_2 + T (N - N_1 - N_2) J \frac{\E \lVert \hat \vtheta_t - \vtheta_t \rVert^2}{\lVert \vtheta_t \rVert} \\
    &\le T N_1 + T N_2 + T N J \frac{\lVert \hatmB_\perp^\top \mB \rVert^2 + k^2 / (\lambda_1^2 N_2)}{\omega} \\
    &\lesssim T N_1 + T N_2 + T N \cdot \frac{d^3 k^2 }{N_1 T}\log^3(NT) + TN \frac{k^2}{N_2} \\
    &\le T N_1 + \frac{N d^3 k^2 }{N_1} \log^3(NT)+ T N_2  + TN \frac{k^2}{N_2} \\
    &\le d^{1.5} k \sqrt{NT} \log^3(NT) + k T \sqrt N. \qedhere
\end{align*}
\end{proof}

\section{Proof of \autoref{thm:inflb}}

In this appendix, we assume all action sets are spherical, i.e. $\mathcal A_{n, t} \equiv \mathbb S^{d-1}$ for $n \in [N], t \in [T]$. Note that these action sets meet \autoref{assum:ellipsoid}. 

For each task $t \in [T]$, we use $R^{N, (t)} = \sum_{n = 1}^N [\max_{\vx \in \gA_{n, t}} \langle \vx, \vtheta_t \rangle - \E{\langle \vx_{n, t}, \vtheta_t \rangle}]$ to denote the regret incurred on task $t$.

\begin{lem} \label{lem:inflb1} Under the setting of \autoref{thm:inflb}, we have $\inf_{\gA} \sup_{\gI} \E[R^{N, T}_{\gA, \gI}] \ge C k T \sqrt{N},$ where $C = 0.0001$.
\end{lem}

\begin{lem} \label{lem:inflb2} 
Under the setting of \autoref{thm:inflb}, we have $\inf_{\gA} \sup_{\gI} \E[R^{N, T}_{\gA, \gI}] \ge C d \sqrt{k T N},$ where $C = 0.0001$.
\end{lem}

\begin{proof}[Proof of \autoref{thm:inflb}] We combine \autoref{lem:inflb1} and \autoref{lem:inflb2}.
\end{proof}

The proofs in this appendix would largely follow the proofs in \autoref{app:lb}, yet this appendix is much longer, because we need to construct instances that satisfies \autoref{assum:ellipsoid}, \autoref{assum:diverse}, and \autoref{assum:w}. 

Our proofs to the lemmas are based on the lower bounds for the (single-task) linearly parameterized bandit setting, which corresponds to the $T = 1$ case in our setting. For this single-task setting, we assume $k = d$ and $\mB = \mI_d$. (This setting need not meet  \autoref{assum:diverse} and \autoref{assum:w}). We write the regret as $R^{N} = R^{N, 1}$ and call algorithms for the single-task setting as single-task algorithms. 

We invoke the following lower bound for the single-task setting. 

\begin{lem} \label{lem:lbsinglegaussian} Assume $N \ge d^2$ and $d \ge 2$. Let $\mathcal N(\mu, \Sigma)$ be the multivariate normal distribution with mean $\mu$ and covariance matrix $\Sigma$. For any single-task algorithm, we have  
\begin{align}
    \E_{\vw \sim \mathcal N(0, \mat I_{d} / d)} [R^N \cdot \ind\{0.09 \le \lVert \vw \rVert \le 3\}] \ge 0.006 d \sqrt N. \label{eq:gauss1}
\end{align}
\end{lem}

The lemma can be proved by following the proof of Theorem 2.1 of \citet{rusmevichientong2010linearly}. Next we use it to prove \autoref{lem:inflb1} and \autoref{lem:inflb2}. The main idea to prove \autoref{lem:inflb1} is to note that we can treat our setting as $T$ independent $k$-dimensional linear bandits. 

Let $\mu_d$ be the conditional probability measure whose density function is given by 
\begin{align*}
    f(x) = \begin{cases} \frac{g(x)}{\Pr_X[0.09 \le \lVert X \rVert \le 3]}, & 0.09 \le \lVert x \rVert \le 3, \\
    0, & \text{otherwise},
    \end{cases}
\end{align*}
where $X \sim \gN(0, \mI_d / d)$ is the multivariate gaussian vector and $g(x)$ is the probability density function of $X$. Then \eqref{eq:gauss1} implies 
\begin{align}
    \E_{w \sim \mu_d } [R^N ] \ge 0.006 d \sqrt N. \label{eq:gauss2}
\end{align}

\begin{proof}[Proof of \autoref{lem:inflb1}] Without loss of generality, we assume $2k$ divides $T$. Suppose, for contradiction, that there is an algorithm $\gA'$, such that for every instance $\gI$, it incurs regret $\E[R^{N, T}_{\gA', \gI}] \le C T k \sqrt{N}$. 
We replace the condition $\lVert w_t \rVert \le 1$ by $\lVert w_t \rVert \le 3$ in our setting. Note that $\gA'$ implies an algorithm $\gA$ that incurs regret $\E[R^{N, T}_{\gA, \gI}] \le 3 C T k \sqrt{N}$.

We construct the following instances $\gI = (\mB, \mW)$, where  $\mB = \mqty(\mI_k & 0)^\top$ and $\mW = \mqty(\vw_1 & \cdots & \vw_T)$ as follows. 
Let 
$$\vw_1 = \cdots \vw_{T / 2k} = \ve_1, \vw_{T / 2k + 1} = \cdots = \vw_{T / k} = \ve_2, \ldots, \vw_{(k-1)T / 2k + 1} = \cdots = \vw_{T / 2} = \ve_k,$$
where $\ve_1, \ldots, \ve_k \in \sR^k$ is the standard basis. Let 
\begin{align*}
    \vw_{T / 2 + 1}, \ldots, \vw_T \sim \mu_d
\end{align*}
be i.i.d. drawn. Thanks to the first $\frac{T}{2}$ tasks, the instance $\mathcal I$ always satisfies \autoref{assum:diverse}. Note that $\lVert \vw_t \rVert \ge 0.09$, so the instance also satisfies \autoref{assum:w}. Then we have 
\begin{align*}
    \E_{\vw_{T / 2 + 1}, \ldots, \vw_T \sim \mu_d}[R^{N, T}_{\gA, \gI}] \le 3 C T k \sqrt N. 
\end{align*}
Thus 
\begin{align*}
    \sum_{t = T / 2 + 1}^T \E_{\vw_{T / 2 + 1}, \ldots, \vw_T \sim \mu_d}[R^{N, (t)}_{\gA, \gI}] \le \E_{\vw_{T / 2 + 1}, \ldots, \vw_T \sim \mu_d}[R^{N, T}_{\gA, \gI}] \le 3 C T k \sqrt N. 
\end{align*}
Therefore, we can find $t \in [T / 2 + 1, T]$ such that 
\begin{align*}
    \E_{\vw_{T / 2 + 1}, \ldots, \vw_T \sim \mu_d}[R^{N, (t)}_{\gA, \gI}] \le \frac{3 CT}{T / 2} k \sqrt N = 6 C k \sqrt{N}.  
\end{align*}
We note that the expectation operator $\E_{w_{T / 2 + 1}, \ldots, w_T \sim \mu_d}$ is over all randomness on tasks $\tau \ne t$ and its parameter $w_{\tau}$. So there is a realization of the randomness on other tasks $\tau \ne t$ that satisfies
\begin{align*}
    \E_{\vw_t \sim \mu_d}[R^{N, (t)}_{\gA, \gI} \mid \vw_{\tau}, \varepsilon_{\tau}, \tau \ne t] \le  6 C k \sqrt N.
\end{align*}

Based on the realization $\vw_\tau, \varepsilon_\tau$, we design a single-task algorithm $\gS$, which plays task $t$ and simulates other tasks $\tau \ne t$ with $w_\tau, \varepsilon_\tau$. The algorithm achieves 
\begin{align*}
    \E_{\vw \sim \mu_d}[R^N_{\gS, \gI}] = \E_{\vw_t \sim \mu_d}[R^{N, (t)}_{\gA, \gI} \mid w_{\tau}, \varepsilon_{\tau}, \tau \ne t] \le 6 C k \sqrt N,
\end{align*} which contradicts to \eqref{eq:gauss2} because $C = 0.0001$. 
\end{proof}

\begin{proof}[Proof of \autoref{lem:inflb2}] Without loss of generality, we assume that $d$ is even and that $2 k$ divides $T$. Suppose, for contradiction, that there is an algorithm $\gA'$, such that for every instance $\gI$, it incurs regret $\E[R^{N, T}_{\gA', \gI}] \le C d \sqrt{k T N}$. 
We replace the condition $\lVert \vw_t \rVert \le 1$ by $\lVert \vw_t \rVert \le 3$ in our setting. Note that $\gA'$ implies an algorithm $\gA$ that incurs regret $\E[R^{N, T}_{\gA, \gI}] \le 3 C  d \sqrt{k T N}$.

We prove the lemma separately, based on whether $k \ge \frac{d}{2}$ or not. 

1. Consider $k \le \frac{d}{2}$. We generate $k$ vectors $\psi_1, \ldots, \psi_k$ so that $\psi_i \sim \mu_{d - i + 1}$. For every dimension $i$, we consider a map 
\begin{align*}
    c_i: \mathbb R^{d \times i} &\to \mathbb R^{d \times (d - i + 1)}, \\ 
    (x_1, \ldots, x_i) &\mapsto (y_1, \ldots, y_{d - i + 1}),
\end{align*}
so that when $\{x_1, \ldots, x_i\}$ are orthogonal, the set $\{y_1, \ldots, y_{d - i + 1}\}$ is the orthonormal basis of the orthogonal complement $\mathrm{span}\{x_1, \ldots, x_i\}^\perp$ of $\mathbb R^d$. Note that $c_i$ can be computed efficiently, e.g. by Gram-Schmidt process. 

We define $\phi_1 = \psi_1$ and $\phi_i = c_{i-1}(\phi_1, \ldots, \phi_{i - 1}) \cdot \psi_i$ for $i \ge 2$. We observe that $\{\phi_1, \ldots, \phi_k\}$ are orthogonal. Next we define our instance $\mathcal I = (\mB, \mat{W})$, where $\mB = \begin{pmatrix} b_1 & \cdots & b_k \end{pmatrix}$ and $b_i = \frac{\phi_i}{\lVert \phi_i \rVert}$. For $\mW = \begin{pmatrix} w_1 & \cdots & w_T \end{pmatrix}$, we let
\begin{align*}
    w_1 = \cdots = w_{T / k} &= \lVert \phi_1\rVert \cdot \ve_1, \\
    w_{T / k + 1} = \cdots = w_{2 T / k} &= \lVert \phi_2 \rVert \cdot \ve_2, \\
    &~~ \vdots \\
    w_{(k-1)T/k + 1} = \cdots = w_T &= \lVert \phi_k \rVert \cdot \ve_k,
\end{align*}
where $\{\ve_1, \ldots, \ve_k\} \subseteq \mathbb R^k$ is the standard basis. Note that the instance $\gI$ always satisfies \autoref{assum:diverse} and \autoref{assum:w} by our choice of $\vw_t$. Note that we have divided the tasks into $k$ groups, so that each group share the same vector $\vw_t$. For $i = 1, \ldots, k$, we denote the regret of group $i$ as 
\begin{align*}
R^{N, ((i))}_{\gA, \gI}   = \sum_{t = (i - 1) T / k + 1}^{i T / k} R^{N, (t)}_{\gA, \gI}.
\end{align*}
For the algorithm $\gA$, it incurs regret 
\begin{align*}
   \E_{\phi_1, \ldots, \phi_k}[R^{N, ((i))}_{\gA, \gI}] \le 3 C d \sqrt{k N T}. 
\end{align*}
So there is a group $i \in [k]$, such that 
\begin{align*}
    \E_{\phi_1, \ldots, \phi_k}[R^{N, ((i))}_{\gA, \gI}] \le \frac{1}{k} 3 C d \sqrt{k N T} = 3 C d \sqrt{\frac{NT}{k}}. 
\end{align*}
Similar to the proof of \autoref{lem:inflb1}, we can fix the randomness for groups $j \ne i$ to obtain a realization, such that
\begin{align*}
    \E_{\phi_i \mid \phi_{j (j \ne i)}}[R^{N, ((i))}_{\gA, \gI} \mid \phi_j, \varepsilon_j, j \ne i] \le \frac{1}{k} 3 C d \sqrt{k N T} = 3 C d \sqrt{\frac{NT}{k}}. 
\end{align*}
Here we note that $\phi_i$ could depend on $\phi_{\iota}$ for $\iota \ge i + 1$. Now we let $\psi'_i \sim \mu_{d - k + 1}$ and let 
\begin{align}
    \phi'_i = A \psi'_i, \qquad A = c_{k - 1}(\phi_1, \ldots, \phi_{i - 1}, \phi_{i + 1}, \ldots, \phi_{k}). \label{eq:lb21}
\end{align} Note that $\phi'_i$ and $\phi_i \mid \phi_{j (j \ne i)}$ are identical, so we have 
\begin{align}
    \E_{\phi_i = \phi'_i}[R^{N, ((i))}_{\gA, \gI} \mid \phi_j, \varepsilon_j, j \ne i] \le \frac{1}{k} 3 C d \sqrt{k N T} = 3 C d \sqrt{\frac{NT}{k}}. \label{eq:lb22}
\end{align}
We complete the proof by showing that \eqref{eq:lb22} implies a single-task algorithm $\gS$ that plays a $(d-k+1)$-dimensional linear  bandit for $\frac{NT }{k}$ times. Let $w = \psi'_i$. Then $w$ is independently drawn from $\mu_{d - k + 1}$. Next we design the algorithm $\gS$, which runs $\gA$ by playing the tasks $t \in \mathcal T = \{ (i - 1) T / k + 1, \ldots, i T / k \}$ and simulates other tasks $\tau \notin \mathcal T$. Note that playing the task $t \in \mathcal T$ is the same as playing the single-task bandit defined by $w$, because we have $\vtheta_t = \mB w_t = \phi_i = A \psi'_i$ and the matrix $A$ is known (as in \eqref{eq:lb21}) after we fix the randomness on other tasks. Since $\lvert \mathcal T \rvert = \frac{T}{k}$ and $\gA$ is played for $N$ times, $\gS$ can play the single-task bandit specified by $w$ for $\frac{NT}{k}$ times. As a result, we have 
\begin{align*}
    \E_{\vw \sim \mu_{d - k + 1}}[R^N_{\gS, \gI}(\mathcal I)] &= \E_{\phi_i = \phi'_i}[R^{N, ((i))}_{\gA, \gI} \mid \phi_j, \varepsilon_j, j \ne i] \\
    &\le 3 C d \sqrt{\frac{NT}{k}} \\
    &\le 6 C (d - k + 1) \sqrt{\frac{NT}{k}},
\end{align*}
which contradicts to \eqref{eq:gauss2} because $C = 0.0001$.

2. Consider $k > \frac{d}{2}$. In this case, the lower bound in \autoref{lem:inflb1} becomes $\Omega(T k \sqrt N) = \Omega(T d \sqrt N)$. Since $T \ge k$, we have $d \sqrt{k T N} \lesssim T d \sqrt N$. Thus we conclude by \autoref{lem:inflb1}. 
\end{proof}

\section{Experiments for Infinite-Arm Setting}
\label{sec:exp_infarm}\paragraph{Setup} In all experiments, we set $d = 10, N = 10^4$ and the action $\mathcal A_t =\mathbb S^{d-1}$. 
The parameters are generated as follows.
We consider $k=2, 3$ in our experiments.
The noise $\varepsilon_{n, t} \sim \mathcal N(0, 1)$ are i.i.d. Gaussian random variables. To verify our theoretical results, we consider a hyper-parameter $c \in \{0.5, 1, 1.5, 2\}$. For each $c$, we run \algoname~ with $N_1 = d^c k \sqrt{\frac{N}{T}}$ and $N_2 = k \sqrt{N}$. 

\paragraph{Results and Discussions}

\begin{figure}[t]
  \centering
  \vspace{-0.5em}
  \resizebox{\textwidth}{!}{\input{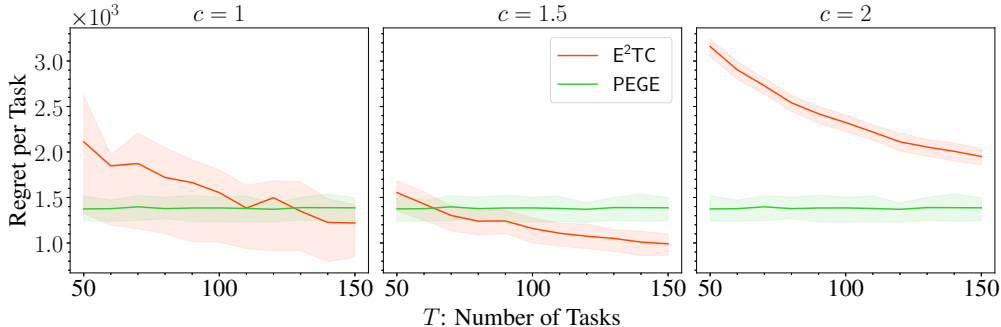}}
  \vspace{-1em}
  \caption{Comparisons of \algoname~with \textsf{PEGE} for $k = 2$.}
  \vspace{-1em}
  \label{fig:infsim2}
\end{figure}

We present the simulation results in \autoref{fig:infsim} and \autoref{fig:infsim2}. We emphasize that the $y$-axis in our figures corresponds to the regret per task, which is defined as $\frac{\mathrm{Reg}_{N, T}}{T}$. 

Our main observation is that only when the number of tasks $T$ is large and we choose the right scaling $N_1 = d^{1.5} k \sqrt{\frac{N}{T}}$, our method can outperform the \textsf{PEGE} algorithm. We discuss several implications of our results. 
First, representation learning does help, especially when there are many tasks available for us to  learn the representation, as we see in all figures that the regret per task of \algoname~ decreases as $T$ increases. 
Second, the help of representation learning is bounded. In the figures, we see that the curves of \algoname~ bends to a horizontal line as $T$ increases, which suggests a lower bound on the regret per task. Meanwhile, we also proved an $\Omega(k \sqrt{N})$ lower bound on the regret per task in \autoref{thm:inflb}. 
Third, representation learning may have adverse effect without enough task. In our figures, this was established by noting that our algorithm cannot outperform \textsf{PEGE} when $T$ is small. This corresponds to the ``negative transfer'' phenomenon observed in previous work \citep{wang2019characterizing}.
Fourth, the correct hyper-parameter $c = 1.5$ is crucial. For hyper-parameter other than $c = 1.5$, the figures show that our algorithm would require much more tasks to outperform \textsf{PEGE}.
Lastly, by comparing the two figures, we notice that our algorithm has bigger advantage when $k$ is smaller, which corroborates the scaling with respect to $k$ in our regret upper bound. In contrast, \textsf{PEGE} does not benefit from a smaller $k$. 

\begin{figure}[t]
  \centering
  \resizebox{\textwidth}{!}{\input{experiments/fig_k33.pgf}}
    \vspace{-1em}
  \caption{Comparisons of \algoname~with \textsf{PEGE} for $k = 3$.}
  \label{fig:infsim}
\end{figure}

\end{document}